\documentclass[11pt]{article}

\PassOptionsToPackage{table}{xcolor}

\usepackage[final]{acl}

\usepackage{times}
\usepackage{latexsym}
\usepackage[T1]{fontenc}
\usepackage[utf8]{inputenc}
\usepackage{microtype}
\usepackage{inconsolata}
\usepackage[dvipsnames]{xcolor}
\usepackage{arydshln}

\usepackage{graphicx}
\usepackage{subcaption}  % companion to caption (loaded by acl.sty); must come after acl

\usepackage{amsmath,amssymb,amsfonts,mathtools}
\usepackage{amsthm}

\usepackage{booktabs}
\usepackage{makecell}
\usepackage{multirow}
\usepackage{float}
\usepackage{placeins}

\usepackage[normalem]{ulem}
\usepackage{enumerate}
\usepackage{enumitem}
\usepackage{tablefootnote}

\usepackage{algorithm}
\usepackage{algorithmic}

\usepackage{tikz}
\usetikzlibrary{positioning, bayesnet}

\theoremstyle{plain}
\newtheorem{theorem}{Theorem}[section]
\newtheorem{proposition}[theorem]{Proposition}
\newtheorem{lemma}[theorem]{Lemma}
\newtheorem{corollary}[theorem]{Corollary}
\theoremstyle{remark}
\newtheorem{remark}[theorem]{Remark}
\theoremstyle{definition}

\title{CGES: Confidence-Guided Early Stopping for Efficient and Accurate Self-Consistency}

\author{%
  Ehsan Aghazadeh\qquad
  Ahmad Ghasemi\qquad
  Hedyeh Beyhaghi\qquad
  Hossein Pishro-Nik \\
  University of Massachusetts Amherst \\
  \texttt{\{eaghazadeh,aghasemi,hbeyhaghi,pishro\}@umass.edu}
}

\begin{document}
\maketitle

\begin{abstract}
Large language models (LLMs) are often queried multiple times at test time, with predictions aggregated by majority vote. While effective, this \emph{self-consistency}~\citep{wang2023selfconsistency} strategy requires a fixed number of calls and fails when the correct answer is infrequent. We introduce \textbf{\emph{\underline{C}onfidence-\underline{G}uided \underline{E}arly \underline{S}topping (CGES)}}, a Bayesian framework that forms posteriors over candidate answers and adaptively halts sampling once one answer accumulates enough posterior mass. 
We prove guarantees in both an ideal calibrated regime and a realistic noisy-confidence regime under a directional drift condition. 
Averaged over five reasoning benchmarks, CGES reduces the average number of calls by \textbf{58\%} on average (from 16.0 to 6.7) while matching its accuracy within \textbf{0.4 percentage points} of self-consistency.
\end{abstract}

\section{Introduction}

Large language models (LLMs) have achieved strong progress across reasoning, problem solving, and open-domain tasks~\citep{singh2026openaigpt5card,comanici2025gemini25pushingfrontier,deepseekai2025deepseekr1incentivizingreasoningcapability,qwen2025qwen25technicalreport}. A common way to improve reliability is \emph{test-time scaling}~\citep{snell2025scaling,wu2025inference,muennighoff2025s1simpletesttimescaling}, a family of methods that allocate additional inference-time computation to improve performance. One subset of these methods samples multiple responses and aggregates them into a final prediction~\citep{wang2023selfconsistency,li2024escape,aggarwal-etal-2023-lets}. Among the most widely used methods, self-consistency (SC) \citep{wang2023selfconsistency} aggregates outputs by majority vote, leveraging the intuition that the most frequent answer across diverse generations is likely to be correct. While simple and effective in many settings, majority-based aggregation suffers from two major shortcomings. First, it assumes that response frequency is a faithful proxy for correctness, which fails in cases where the correct answer appears infrequently~\citep{huang-etal-2024-mirror,taubenfeld-etal-2025-confidence}. Second, it requires a fixed number of model calls regardless of confidence, leading to substantial inefficiency~\citep{li2024escape,aggarwal-etal-2023-lets,wang-etal-2025-make}.

Confidence signals offer an alternative perspective. Instead of depending solely on frequency, one can incorporate confidence scores that capture the model's belief in each response~\citep{kadavath2022languagemodelsmostlyknow,geng-etal-2024-survey}. These scores may be derived from different sources. \emph{Token probabilities} are taken directly from the model's output distribution and reflect how certain the model is about generating each token~\citep{malinin2021uncertainty,bakman-etal-2024-mars}. \emph{Calibration schemes} adjust these raw probabilities so that they better match the actual likelihood of correctness, turning overconfident or underconfident estimates into more reliable signals~\citep{kadavath2022languagemodelsmostlyknow,yaldiz-etal-2025-design}. \emph{External reward models} are trained separately, often with human feedback or domain-specific supervision, and provide an independent measure of response quality beyond the model's own probabilities~\citep{lightman2024lets,zhang2025lessonsdevelopingprocessreward}. Such signals can distinguish between frequent but uncertain answers and rare yet confident ones. 
This makes confidence a direct remedy for the central failure mode of self-consistency: majority frequency can dominate even when the minority answer carries stronger evidence.

\begin{figure*}[t]
  \centering
  \includegraphics[width=0.5\linewidth]{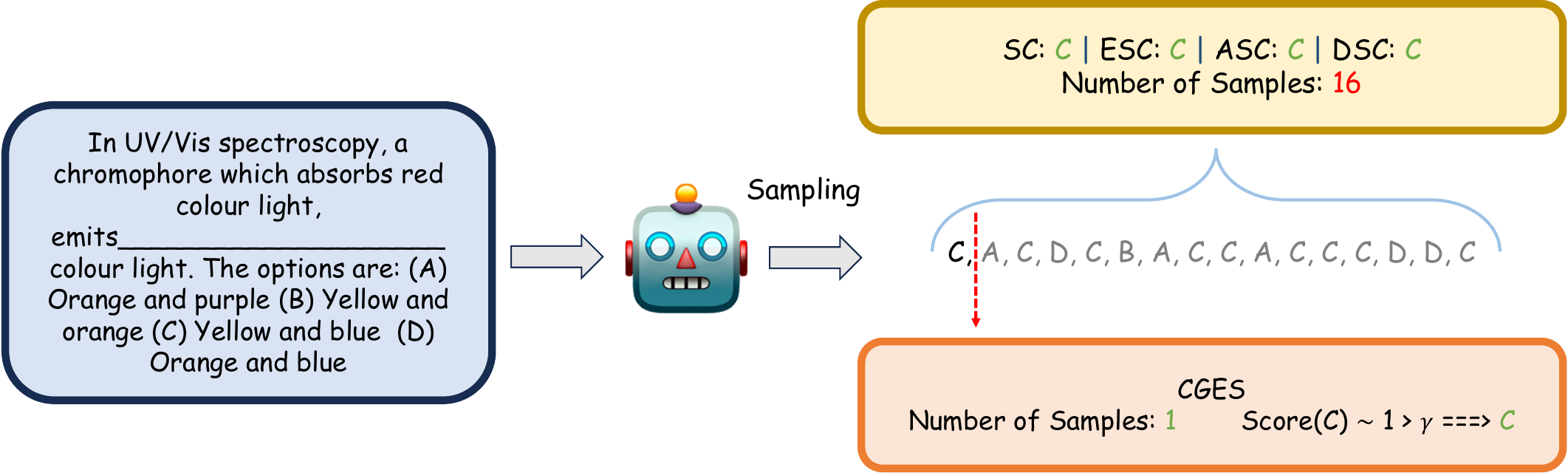}\\[-1pt]
  {\tiny\emph{(a) Efficiency: all baselines reach the correct answer (C) but use the full budget of 16 samples, while CGES stops after one sample once Score(C)$\,\sim\,1>\gamma$.}}\\[3pt]
  \includegraphics[width=0.55\linewidth]{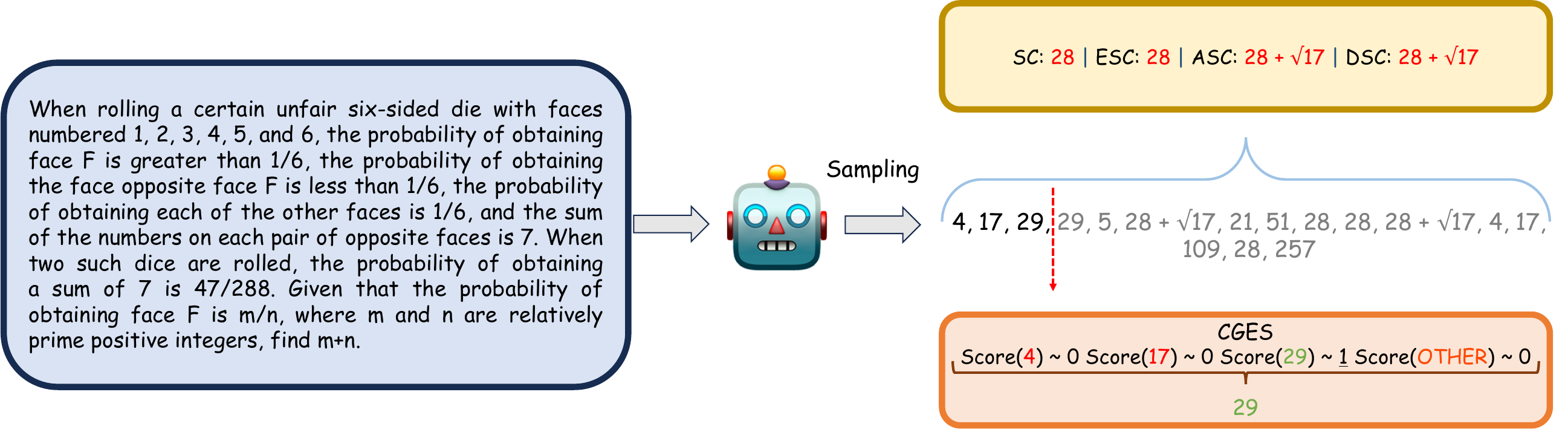}\\[-1pt]
  {\tiny\emph{(b) Accuracy: SC/ESC return the plurality answer 28 and ASC/DSC return $28{+}\sqrt{17}$, both wrong; CGES aggregates confidences and recovers the correct minority answer 29.}}
  \caption{\textbf{Examples of CGES vs.\ baselines.} Top: confidence-driven early stopping saves samples when the model is already certain. Bottom: confidence-weighted aggregation recovers a minority-but-confident answer that majority voting misses.}
  \label{fig:motivating}
\end{figure*}

We propose a \emph{confidence-based Bayesian framework} for self-consistency. Rather than counting answer frequencies, our approach treats each sampled response and its confidence as probabilistic evidence, yielding posterior-style scores over candidate answers. 
This gives CGES two advantages over majority voting: (1) when the correct answer is frequent, CGES often reaches the same decision with fewer samples by stopping once posterior mass concentrates; and (2) when the correct answer is a minority, CGES can still recover it if correct samples have stronger confidence than frequent but uncertain alternatives. Figure~\ref{fig:motivating} illustrates both cases with real examples. 

\textbf{We formalize this idea under both an \emph{ideal} calibrated-confidence regime and a \emph{realistic} noisy-confidence regime that reflects practical LLM inference, where confidence estimates may be imperfect and only directionally informative.} \textbf{The realistic analysis shows that CGES remains consistent as long as confidence signals are even weakly informative about the correct answer, and our experiments validate this behavior with practical confidence estimators.}

To operationalize this framework, we introduce \emph{Confidence-Guided Early Stopping (CGES)}, which combines Bayesian scoring with adaptive stopping. CGES trades off accuracy and efficiency by stopping once posterior concentration exceeds a threshold or the budget is reached. We evaluate CGES on AIME24, MATH500~\citep{hendrycks2021measuringa}, GSM8K~\citep{cobbe2021trainingverifierssolvemath}, GPQA~\citep{rein2024gpqa}, and MMLU\_Pro~\citep{wang2024mmlupro}, comparing against self-consistency~\citep{wang2023selfconsistency} and early-stopping self-consistency~\citep{li2024escape}. CGES consistently reduces LLM calls by large margins while maintaining or improving accuracy, showing the value of calibrated confidence for test-time scaling and principled Bayesian aggregation. Our implementation is publicly available.\footnote{\url{https://github.com/EhsanAghazadeh/cges}}

Unlike prior self-consistency methods that separate \textbf{aggregation} (e.g., majority vote) from \textbf{stopping} (e.g., agreement-based heuristics), our approach casts both within a single probabilistic framework. We treat sampled responses together with their confidence scores as evidence for posterior inference over candidate answers, and use posterior concentration to drive adaptive stopping. This yields a unified view of test-time scaling in which confidence is not merely an auxiliary signal, but a mechanism for both more principled aggregation and more efficient inference.

Our contributions are as follows:
\vspace{-5pt}
\begin{itemize}[leftmargin=1.2em, itemsep=2pt, parsep=0pt]
    \item We reformulate self-consistency as \emph{posterior inference over candidate answers}, replacing frequency-based majority voting with confidence-aware Bayesian aggregation.

    \item We introduce \emph{Confidence-Guided Early Stopping (CGES)}, a unified framework in which the same posterior scores are used both to aggregate sampled responses and to decide when further sampling is unnecessary.

    \item We establish guarantees under an \textit{ideal} calibrated-confidence setting in Theorem~\ref{thrm:ideal}, and, more importantly, under a \textit{realistic} noisy-confidence regime in Theorem~\ref{thm:realistic}, where ideal assumptions are relaxed. \textbf{We show that CGES remains effective beyond idealized assumptions.}

    \item We empirically validate CGES across five reasoning benchmarks, showing that confidence-aware aggregation can substantially reduce inference cost while maintaining or improving accuracy relative to standard self-consistency and heuristic early-stopping baselines.
\end{itemize}

\section{Related Work}

A widely used test-time scaling method is \emph{self-consistency}~\citep{wang2023selfconsistency}, which aggregates multiple reasoning paths by majority vote. While effective, it uses a fixed number of calls and can fail when the correct answer is infrequent~\citep{huang-etal-2024-mirror,taubenfeld-etal-2025-confidence}. Several variants improve efficiency via dynamic stopping rules~\citep{li2024escape,aggarwal-etal-2023-lets,wang-etal-2025-make,huang2025efficienttesttimescalingselfcalibration}. Recent methods also use confidence-weighted aggregation~\citep{taubenfeld-etal-2025-confidence} or exploit minority responses~\citep{huang-etal-2024-mirror}. Unlike prior work, we introduce a \textbf{Bayesian framework} with \textbf{theoretical guarantees} for \textbf{confidence-guided early stopping}, unifying aggregation and adaptive stopping in one probabilistic model. Concurrent work also studies confidence in test-time scaling. DeepConf~\citep{fu2026deep} uses token-level confidence within a single trajectory to halt low-confidence generations. ARROL~\citep{xu2026prunegenerateonlinerollout} learns a quality head during RLVR training and reuses it as voting weights at inference. Both are complementary to CGES, a training-free Bayesian rule for aggregation and stopping over completed responses, which can directly use such learned confidence signals.

Beyond self-consistency, a broader literature studies adaptive allocation of test-time compute. One line examines compute-optimal scaling and inference scaling laws, proposing budget-aware sampling strategies~\citep{snell2025scaling,wu2025inference,wang2025every,muennighoff2025s1simpletesttimescaling}. Another combines search with verification, using tree-search expansions, step-wise verifiers, iterative elimination, and query-variant ensembling to prune or refine candidates~\citep{bi2025forestofthought,lample2022hypertreeproofsearchneural,koh2024treesearchlanguagemodel,li-etal-2023-making,lightman2024lets,chen2026provable,huang2024dividereweightconquerlogit}.
These methods typically rely on structured search, verifier signals, or semantic clustering to select a single best reasoning path. Our approach is complementary: it treats confidence as probabilistic evidence and \emph{aggregates} sampled responses through a lightweight Bayesian update with formal consistency guarantees.

A core component of our framework is \emph{confidence estimation}, studied extensively for generative LLMs. Unlike uncertainty quantification, which characterizes the predictive distribution, confidence estimation is response-specific, reflecting the model's belief in a generated output's correctness.~\citep{lin2024generating,kadavath2022languagemodelsmostlyknow}. Existing methods derive confidence from token probabilities with length normalization or learned scoring~\citep{malinin2021uncertainty,bakman-etal-2024-mars,yaldiz-etal-2025-design}, information-theoretic or similarity-based black-box signals~\citep{abbasi-yadkori2024to,bhattacharjya-etal-2025-simba}, and Bayesian or distillation-based estimators~\citep{vejendla2025efficientuncertaintyestimationdistillation}, with surveys and taxonomies providing broader context~\citep{geng-etal-2024-survey,10.5555/3762387.3762592}. Although CGES requires a confidence signal, our goal is not to introduce a new estimator. Instead, we adapt existing techniques within CGES, modifying them when needed; when response-level scores are unavailable, uncertainty quantification methods can serve as confidence proxies.

\section{Confidence-Based Approach}

\subsection{Problem Setting and Notations}
\label{notations}

Suppose we query a large language model (LLM) multiple times with a given query $Q$, whose true answer is $A$, thereby obtaining a sample set $\mathcal{S}$ of responses.
Let $\mathcal{U} = \{a_1, a_2, \dots, a_K\}$ denote the complete set of all possible distinct candidate final answers produced by the LLM, augmented with an explicit \emph{OTHER/unseen} category. The parameter $K$ is the maximum number of such candidates.\footnote{Since the answer space includes at least one concrete candidate and the \emph{OTHER/unseen} category, we necessarily have $K \ge 2$.}

For the theoretical analysis here, we condition on the true answer $A$ appearing in the candidate set, i.e., there exists $I \in [K]$ with $a_I = A$, and use $I$ to denote the unknown correct index. If $A \notin \mathcal{U}$, no method operating only on $\{a_1,\dots,a_K\}$ can recover $A$. In such cases, CGES still returns the highest-posterior candidate in $\mathcal{U}$, but the consistency guarantees in Sections~\ref{sec:ideal-theory}--\ref{sec:realistic-theory} do not apply.

Conditioned on $(Q,A)$ and the identity of the correct candidate $I$, the LLM induces an unknown but fixed response distribution $\textbf{P} = (P_1,\dots,P_K)$ on $\mathcal{U}$, with $P_j = \mathbb{P}[R = a_j \mid Q,A,I]$. Intuitively, this captures the stochastic behavior of the LLM on the query: the likelihood of producing each candidate answer 
depends both on the query and on which candidate is correct. For each LLM call $t=1,2,\dots$, we draw a response--confidence pair $(R_t, C_t) \sim f_{R,C\mid \mathbf{P}}(\cdot,\cdot \mid \mathbf{P})$ from a joint conditional distribution, with marginal $R_t \sim \mathbf{P}$. In general $C_t$ may depend on $R_t$, since it is computed from the trajectory that produced it.
\begin{figure}[t]
\centering
\scalebox{0.6}{
\begin{tikzpicture}
    \node[latent] (Q) {$Q,A$};
    \node[latent, right=1.9cm of Q] (I) {$I$};
    \node[latent, below right=1.0cm and 1.0cm of Q] (P) {$P$};
    \node[obs, right=1.9cm of P, yshift=0.8cm] (R) {$R_t$};
    \node[obs, right=1.9cm of P, yshift=-0.8cm] (C) {$C_t$};
    \edge {Q,I} {P};
    \edge {P} {R,C};
    \edge {R} {C};
    \plate {T} {(R)(C)} {$t=1{:}m$};
\end{tikzpicture}
}
\caption{Graphical model for the LLM sampling process. The edge $R_t \to C_t$ reflects that the confidence score is computed from the trajectory that produced the response, e.g., token probabilities or a reward-model score on $R_t$.}
\label{fig:gm}
\end{figure}
The ideal Bayesian analysis in Theorem~\ref{thrm:ideal} uses only the conditional likelihood $\mathbb{P}(R_t \mid C_t, I)$ specified in Assumption~\ref{assump:ideal}, and therefore does not assume independence between $R_t$ and $C_t$.
Here $R_t$ denotes the \emph{final textual answer string} produced by the LLM on call $t$ (e.g., ``12'', ``$x=3$'', ``yes''), after any post-processing or extraction step. Thus $R_t$ takes values in the finite candidate set $\mathcal{U}$, and is not the full sequence of generated tokens. The confidence signal $C_t$ is a scalar attached to $R_t$ and serves as a (noisy) proxy for its probability of correctness $\textbf{P}$.
This structure can be represented as a graphical model, illustrated in Figure~\ref{fig:gm}.

\paragraph{Idealistic vs.\ Realistic Assumptions.}
We distinguish two roles for our assumptions. \emph{Idealistic} assumptions~\ref{assump:ideal}--\ref{assump:conf} are simplifying conditions used to \emph{derive} the clean Bayesian scoring rule of Section~\ref{subsec:algo-score} and the optimality result of Theorem~\ref{thrm:ideal}; under~\ref{assump:ideal}, each $C_t$ is treated as the \emph{exact} conditional correctness probability of $R_t$, encoded in Eq.~\eqref{eq:ideal_conditional}. \emph{Realistic} assumptions~\ref{assump:independence}--\ref{assump:uniform} are mild and expected to hold in practice. The consistency guarantee under realistic data (Theorem~\ref{thm:realistic}) is proved using \emph{only}~\ref{assump:independence} and~\ref{assump:uniform}; it does not require~\ref{assump:ideal} or~\ref{assump:conf}. In this regime $C_t$ is a noisy estimator of correctness (e.g., a token-probability score or reward-model output) and may depend on $R_t$. Section~\ref{experiments} shows that CGES remains effective on real LLM data, where the idealistic assumptions do not hold.

The four assumptions are as follows:
\vspace{-7pt}
\begin{enumerate}[leftmargin=1.2em, itemsep=2pt, parsep=0pt]
    \item \label{assump:independence} Given a fixed query $Q$ and correct index $I$ (and hence fixed $\mathcal{U}$ and $\mathbf{P}$), the samples $\{(R_t,C_t)\}_{t=1}^m$ are i.i.d. This is standard in self-consistency and test-time scaling analyses~\citep{snell2025scaling,wu2025inference}.
    
    \item \label{assump:uniform}
    $I \sim \text{Uniform}(\{1,\dots,K\})$. This prior over the correct candidate's \emph{index} treats all $K$ hypotheses symmetrically:
    the labeling of $\mathcal{U}$ is arbitrary, so the correct candidate's position carries no information and does not affect the proposed method's performance.

    \item \label{assump:ideal} Under hypothesis $I = i$, the correct answer $a_i$ is emitted with probability $C_t$, while each incorrect candidate shares the residual probability mass uniformly:
    \begin{equation}
        \begin{aligned}
            \mathbb{P}(R_t = a_i \mid C_t, I=i) &= C_t, \\
            \mathbb{P}(R_t = a_j \neq a_i \mid C_t, I=i) &= \frac{1-C_t}{K-1}.
        \end{aligned}
    \end{equation}

    \item \label{assump:conf} The confidence score $C_t$ is independent of the index of the correct answer: $\mathbb{P}(C_t\mid I=i) = \mathbb{P}(C_t\mid I=j)$ for all $i,j \in\{1,\dots,K\}$. This is a labeling-symmetry condition on the \emph{marginal} of $C_t$ and does \emph{not} rule out larger $C_t$ on correct answers.
\end{enumerate}

Given these assumptions, the objectives of our framework are twofold:
(i) to \emph{identify the most probable index $i \in [K]$ corresponding to the true answer}, and
(ii) to \emph{quantify the level of confidence in this selection}.

\subsection{Bayesian Confidence-Based Framework}
\label{framework}

We infer the (unknown) index of the correct candidate, denoted by
$I \in \{1,\dots,K\}$, from observed response--confidence pairs
\(
\mathrm{Obs} \triangleq \{(R_t,C_t)\}_{t=1}^m.
\)

\begin{lemma}[Posterior simplification under confidence-independence]
\label{lem:posterior_simplification}
Assume: (i) the samples $\{(R_t,C_t)\}_{t=1}^m$ are i.i.d.\ (Assumption~\ref{assump:independence}), (ii) the prior over indices is uniform, $\mathbb{P}(I=i)=1/K$ (Assumption~\ref{assump:uniform}), and (iii) the confidence score is independent of the correct index, i.e.,
$\mathbb{P}(C_t\mid I=i)=\mathbb{P}(C_t\mid I=j)$ for all $i,j$ (Assumption~\ref{assump:conf}).
Then the posterior over $I$ satisfies
\begin{equation}
\label{eq:pos}
\mathbb{P}(I{=}i \mid \mathrm{Obs})
\;=\;
\frac{\prod_{t} \mathbb{P}(R_t \mid C_t, I{=}i)}
     {\sum_{k=1}^K \prod_{t} \mathbb{P}(R_t \mid C_t, I{=}k)}.
\end{equation}
In particular, the marginal law of $\{C_t\}_{t=1}^m$ does not affect the posterior beyond the conditioning in $\mathbb{P}(R_t \mid C_t, I=i)$.
\end{lemma}

\noindent\textbf{Proof Sketch.}
Bayes' rule combined with the uniform prior (Assumption~\ref{assump:uniform}) reduces the posterior to a normalized product of per-sample joint likelihoods; the chain rule decomposes each joint, and Assumption~\ref{assump:conf} cancels the resulting $\mathbb{P}(C_t)$ factor common to numerator and denominator, yielding~\eqref{eq:pos}. A full proof is provided in Appendix~\ref{appendix:lemma_proof}.

\noindent
\textbf{Model instantiations.}
Lemma~\ref{lem:posterior_simplification} reduces posterior inference to specifying the conditional model
$\mathbb{P}(R_t\mid C_t,I=i)$, which under ideal calibration
(Assumption~\ref{assump:ideal}) has the explicit form:
\begin{equation}
\label{eq:ideal_conditional}
\mathbb{P}(R_t \mid C_t, I=i)=
\begin{cases}
C_t & \text{if } R_t=a_i, \\[6pt]
\frac{1-C_t}{K-1} & \text{otherwise},
\end{cases}
\end{equation}
which is the model used in Theorem~\ref{thrm:ideal}.
In the \emph{realistic} setting of Theorem~\ref{thm:realistic}, Assumptions~\ref{assump:ideal} and~\ref{assump:conf} are dropped: $C_t$ is a noisy estimator of correctness that may depend on $R_t$, and the formula in Eq.~\eqref{eq:pos} is used as a \emph{scoring rule} rather than as a true posterior. Theorem~\ref{thm:realistic} establishes consistency of this scoring rule under only Assumptions~\ref{assump:independence} and~\ref{assump:uniform}.

\subsection{Algorithm for Confidence-Based Scoring}
\label{subsec:algo-score}
Given sampled answers and their confidences \(\mathcal{S}=\{(R_t,C_t)\}_{t=1}^{m}\) for a single query, our goal is to convert them into normalized, confidence-weighted scores over the candidate set $\mathcal{U} = \{a_1, a_2, \dots, a_K\}$. 
Under Assumptions~\ref{assump:independence}--~\ref{assump:conf}, these scores coincide with Bayesian posterior probabilities; otherwise, they are pseudo-posteriors with consistency characterized in Theorem~\ref{thm:realistic}. 
The scoring rule follows from the factorized posterior in Eq.~\eqref{eq:pos}: for each candidate \(a_i\), we test the hypothesis that \(a_i\) is the correct answer. 
Under this hypothesis, each observation \((R_t,C_t)\) contributes \(C_t\) when \(R_t=a_i\), and \((1-C_t)/(K-1)\) otherwise. 
Thus, the unnormalized score for \(a_i\) is 
\(s_{a_i}=\prod_{t:R_t=a_i} C_t \prod_{t:R_t\neq a_i} (1-C_t)/(K-1)\), 
which is normalized as 
\(\mathrm{score}(a_i)=s_{a_i}/\sum_{j=1}^{K}s_{a_j}\). 
Lines 1--5 of Algorithm~\ref{alg:cges} implement this \textsc{Score} routine.

Lines 6 through 17 wrap \textsc{Score} into an adaptive loop that allocates test-time compute per question. We begin with one sample per question, compute posteriors via \textsc{Score}, and maintain the set of unresolved questions \(D_{\text{rem}}\) whose current top posterior is below a confidence threshold \(\gamma\). At each subsequent round, we query the LLM only for questions in \(D_{\text{rem}}\), append the new \((R_t^n,C_t^n)\), recompute \textsc{Score} on that question's accumulated samples, and remove the question from \(D_{\text{rem}}\) once its top posterior exceeds \(\gamma\). The process halts when all questions are confident or the budget \(B\) is reached, returning the argmax label per question along with the average number of LLM calls. If no candidate ever exceeds \(\gamma\) before the budget is exhausted, CGES still returns the current argmax label, so \(\gamma\) controls test-time compute rather than enforcing abstention. Extending the framework with a no-answer option is left for future work.

The theoretical analysis in Sections~\ref{sec:ideal-theory} and \ref{sec:realistic-theory} treats $\mathcal{U}$ and $K \ge 2$ as fixed. In the online implementation, we instantiate $\mathcal{U}_t$ as the answers observed through round $t$ together with an explicit OTHER or unseen bucket, ensuring $K_t \ge 2$ from the first round onward. Consistency extends to this dynamic-support case (Appendix~\ref{app:dynamic-support}).

\begin{algorithm}[t]
  \caption{\small Confidence-Guided Early Stopping (CGES)}
  \label{alg:cges}
  \scriptsize
  \begin{algorithmic}[1]
    \REQUIRE Candidate set $\mathcal{U}=\{a_1,\dots,a_K\}$; $N$ questions; threshold $\gamma$; budget $B$
    \STATE \textbf{function} \textsc{Score}($\mathcal{S}=\{(R_t,C_t)\}_{t=1}^m$)
    \STATE \quad \textbf{for} $a_i \in \mathcal{U}$ \textbf{do} $s_{a_i}\gets\prod_{t\,;\,R_t=a_i}C_t\,\cdot\,\prod_{t\,;\,R_t\neq a_i}\tfrac{1-C_t}{K-1}$
    \STATE \quad $Z \gets \sum_{a_i} s_{a_i}$
    \STATE \quad \textbf{return} $\{s_{a_i}/Z\}_{a_i \in \mathcal{U}}$
    \STATE \textbf{end function}
    \STATE Initialize $\mathrm{scores}_n \gets \textsc{Score}(\{(R_1^n,C_1^n)\})$ for all $n\in[N]$; set $D_{\text{rem}}\gets[N]$, $\mathrm{calls}\gets N$
    \FOR{$t = 2$ to $B$}
      \STATE $D_{\text{rem}}\gets\{n\in D_{\text{rem}}\,;\,\max_i \mathrm{scores}_n[i] < \gamma\}$
      \IF{$D_{\text{rem}}=\emptyset$} \STATE \textbf{break} \ENDIF
      \STATE Query LLM for each $n\in D_{\text{rem}}$; $\mathrm{calls}\mathrel{+}=|D_{\text{rem}}|$
      \FOR{$n\in D_{\text{rem}}$}
        \STATE $\mathrm{scores}_n \gets \textsc{Score}(\{(R_1^n,C_1^n),\dots,(R_t^n,C_t^n)\})$
      \ENDFOR
    \ENDFOR
    \STATE \textbf{return} $\hat{y}_n = a_{\arg\max_i \mathrm{scores}_n[i]}$ for all $n$, and $\mathrm{usage}=\mathrm{calls}/N$
  \end{algorithmic}
\end{algorithm}

\subsection{Theoretical Analysis: Ideal Scenario}
\label{sec:ideal-theory}

We now analyze CGES under an ideal calibrated-confidence model. 
The following theorem shows that the CGES posterior concentrates on the true answer as \(m\) grows.

\begin{theorem}
\label{thrm:ideal}
Under Assumptions~\ref{assump:independence}, \ref{assump:uniform}, \ref{assump:ideal}, and \ref{assump:conf}, and provided $C_t \in (0,1)$ almost surely with $\mathbb{P}(C_t = 1/K) = 0$ and $\mathbb{E}\!\left[\left|\log\!\tfrac{C_t(K-1)}{1-C_t}\right|\right] < \infty$, the Bayesian confidence-based aggregator identifies the correct answer with probability tending to one as the number of samples $m$ grows:
\begin{equation*}
\mathbb{P}\!\left(\arg\max_{i\in[K]} X_i = I\right)
\longrightarrow 1 \quad \text{as} \quad m\to\infty,
\end{equation*}
where
$X_i = \prod_{t=1}^{m}\mathbb{P}(R_t\mid C_t,I=i) \,\big/\, \sum_{k=1}^K \prod_{t=1}^{m}\mathbb{P}(R_t\mid C_t,I=k)$.
In fact, $X_I\to 1$ and $X_k\to 0$ for all $k\neq I$ almost surely.
\end{theorem}

\noindent\textbf{Proof Sketch.}
For any wrong index $k \neq I$, the log-likelihood ratio of the samples under hypotheses $I$ vs.\ $k$ has strictly positive expected increment whenever $C_t \neq 1/K$. The Strong Law of Large Numbers then drives the cumulative ratio to $+\infty$, forcing $X_I \to 1$ and $X_k \to 0$ a.s. Full proof in Appendix~\ref{appendix:ideal_full_proof}.

Theorems~\ref{thrm:ideal} and~\ref{thm:realistic} characterize the limiting behavior of $X_i$ as the number of samples grows. The next proposition gives the CGES threshold $\gamma$ a calibrated meaning at the actual (finite) stopping time used by Algorithm~\ref{alg:cges}.

\begin{proposition}[Posterior-risk control under early stopping]
\label{prop:risk_bound}
Let Assumptions~\ref{assump:independence}--\ref{assump:conf} hold and fix $\gamma \in (0,1)$. Define the threshold-stopping time $\tau_\gamma := \inf\{m \ge 1 : \max_i X_i^{(m)} \ge \gamma\}$ and the returned label $\hat{I} := \arg\max_i X_i^{(\tau_\gamma)}$ (with ties broken arbitrarily). On the event $\{\tau_\gamma \le B\}$, $\mathbb{P}\!\bigl(I \ne \hat{I} \,\bigm|\, \mathrm{Obs}_{\tau_\gamma}\bigr) \;\le\; 1 - \gamma.$
\end{proposition}

\noindent\textbf{Proof Sketch.}
Lemma~\ref{lem:posterior_simplification} combined with Eq.~\eqref{eq:ideal_conditional} identifies the CGES score with the true Bayesian posterior at every sample size $m$. The stopping rule fires precisely when $\max_i X_i^{(\tau_\gamma)} \ge \gamma$, so the posterior on the returned candidate is at least $\gamma$, and its complement gives the stated risk bound. A complete proof is provided in Appendix~\ref{appendix:risk_bound_proof}.

In fact, no decision rule that uses only the same response--confidence observations attains a smaller conditional probability of error than CGES at $\tau_\gamma$, and $\tau_\gamma$ is the earliest sample count at which any such rule could have certified the same posterior-risk level (rigorous statement and proof in Appendix~\ref{appendix:risk_bound_proof}, Corollary~\ref{cor:optimality}).

\begin{remark}[Tightness--frequency tradeoff at $\gamma \to 1$]
\label{rem:gamma_tradeoff}
The bound trades tightness against frequency of triggering: as $\gamma \to 1$, the threshold-stopping event $\{\tau_\gamma \le B\}$ becomes rare, and has probability zero at $\gamma = 1$ since $C_t \in (0,1)$ a.s.\ implies $\max_i X_i^{(m)} < 1$ for every finite $m$. The bound is informative for moderate $\gamma$, where $1-\gamma$ is small and the algorithm halts with positive probability before exhausting the budget.
\end{remark}

In the realistic regime of Theorem~\ref{thm:realistic}, $X_i$ is a pseudo-posterior rather than the true posterior, so Proposition~\ref{prop:risk_bound} becomes a pseudo-risk statement; empirically (Section~\ref{experiments}), $\gamma$ continues to provide a useful efficiency--accuracy knob.

\subsection{Theoretical Analysis: Noisy Confidence Signals}
\label{sec:realistic-theory}
We now drop the two idealistic Assumptions~\ref{assump:ideal} and~\ref{assump:conf} and ask whether CGES remains consistent when confidence is only a noisy, possibly response-dependent signal.

\begin{theorem}[Consistency under noisy confidence signals]
\label{thm:realistic}
Let $\ell(c) := \log\!\frac{c(K-1)}{1-c}$. Assume $\{(R_t,C_t)\}_{t=1}^m$ are i.i.d.\ samples from an arbitrary joint distribution over $\mathcal{U}\times(0,1)$, with $\mathbb{E}\!\left[|\ell(C_t)|\right] < \infty$. Let $X_i$ denote the normalized score computed by CGES (Algorithm~\ref{alg:cges}) 
using the auxiliary likelihood of Eq.~\eqref{eq:ideal_conditional}: under all four Assumptions~\ref{assump:independence}--\ref{assump:conf}, $X_i$ coincides with the true Bayesian posterior on candidate $a_i$ via Lemma~\ref{lem:posterior_simplification}; under only Assumptions~\ref{assump:independence}--\ref{assump:uniform}, $X_i$ is a normalized scoring rule that need not equal the true posterior. Without loss of generality, let $a_1$ denote the correct answer.

For each incorrect candidate $a_k$ ($k\neq 1$), define the expected log-likelihood drift
\[
\mu_k
\;:=\;
\mathbb{E}\!\left[
\bigl(\mathbf{1}\{R_t = a_1\} - \mathbf{1}\{R_t = a_k\}\bigr)\,\ell(C_t)
\right],
\]
where the expectation is over the true joint distribution of $(R_t, C_t)$.

If $\mu_k > 0$ for every $k \neq 1$, then $X_1 \xrightarrow{\mathrm{a.s.}} 1$
and
$X_k \xrightarrow{\mathrm{a.s.}} 0$
for all $k \neq 1$,
as $m \to \infty$.
Conversely, if $\mu_{k^\star} < 0$ for some $k^\star \neq 1$, then $X_1 \xrightarrow{\mathrm{a.s.}} 0$
as $m \to \infty$.
\end{theorem}

\paragraph{Interpretation.}
The drift $\mu_k$ is the gap in expected scoring reward $\ell(C_t)$ between emissions of the true answer $a_1$ and emissions of $a_k$, under the true joint of $(R_t, C_t)$. The condition $\mu_k > 0$ requires neither calibrated confidences nor any independence between $R_t$ and $C_t$; it allows even $\mathbb{P}(R_t = a_1) < \mathbb{P}(R_t = a_k)$ when correct emissions carry sufficiently higher confidence. This is the minority-but-confident regime in which majority voting fails (Figure~\ref{fig:motivating}b). The boundary $\mu_k = 0$ is not covered. \textbf{The auxiliary scoring rule derived from the ideal model retains its consistency under the realistic assumptions alone whenever $\mu_k > 0$,} a condition automatic under the ideal model. Only under the ideal model does $\gamma$ carry a calibrated posterior-risk interpretation (Proposition~\ref{prop:risk_bound}); in the noisy regime it acts as an empirical accuracy--efficiency knob (Section~\ref{experiments}).

\vspace{3pt}
\noindent\textbf{Proof Sketch.}
For any $k \neq 1$, the log-likelihood ratio of the samples under hypotheses $I=1$ vs.\ $I=k$ has one-step expected increment $\mu_k$ under the true joint. If $\mu_k > 0$, the Strong Law of Large Numbers forces the cumulative ratio to $+\infty$ and $X_1 \to 1$ a.s.; if $\mu_{k^\star} < 0$ for some $k^\star$, the cumulative ratio diverges to $-\infty$ and $X_1 \to 0$ a.s. Full proof in Appendix~\ref{appendix:real_full_proof}.

\vspace{3pt}

\noindent
\textbf{Dynamic candidate set.}
Theorems~\ref{thrm:ideal} and~\ref{thm:realistic} fix the candidate set $\mathcal{U}$ and its size $K$, whereas the online implementation grows them with the observed answers (Section~\ref{subsec:algo-score}). The consistency guarantee extends to this dynamic-support version: once the observed support stabilizes, the positive-drift condition of Theorem~\ref{thm:realistic} drives the true answer's normalized score to one (rigorous statement and proof in Appendix~\ref{app:dynamic-support}, Proposition~\ref{prop:dynamic-support-stabilization}). The result is asymptotic and does not preclude premature stopping before the true answer first appears.

\begin{remark}[Missing ground-truth among candidates]\label{remark:Missing}
Theorems~\ref{thrm:ideal} and~\ref{thm:realistic} analyze CGES under the event \(A\in\mathcal{U}\), so \(I\) is well-defined.
If \(A\notin\mathcal{U}\), any procedure restricted to \(\mathcal{U}\) cannot return \(A\): the error probability has an irreducible term \(\mathbb{P}(A\notin\mathcal{U})\) that no self-consistency-style scheme can remove.
In this misspecified regime, CGES (like other sampling-based methods) selects the candidate in \(\mathcal{U}\) with the highest posterior mass, and the above consistency guarantees do not apply.
In practice, a high threshold \(\gamma\) can flag such cases (e.g., when no candidate attains substantial posterior probability), but correctness guarantees are impossible without access to \(A\).
\end{remark}

\section{Experiments}
\label{experiments}
We evaluate CGES on five reasoning benchmarks with two open-source 7B models. The experiments answer three questions. \textbf{Does confidence-aware aggregation match self-consistency accuracy with fewer model calls? Do the gains survive wall-clock timing and matched-budget comparisons? Does the realistic-regime theory in Theorem~\ref{thm:realistic} predict where CGES wins and where it does not?}

\subsection{Setup}
\label{subsec:setup}
We use \textbf{AIME24} (30 problems), \textbf{MATH500} \citep{hendrycks2021measuringa}, \textbf{GSM8K}~\citep{cobbe2021trainingverifierssolvemath}, \textbf{MMLU\_Pro}~\citep{wang2024mmlupro}, and \textbf{GPQA Diamond}~\citep{rein2024gpqa}. Harder tasks (AIME24, GPQA) use \emph{DeepSeek-R1-Distill-Qwen-7B}~\citep{deepseekai2025deepseekr1incentivizingreasoningcapability}; the rest use \emph{Qwen2.5-7B}~\citep{qwen2025qwen25technicalreport}. Baselines are SC~\citep{wang2023selfconsistency}, ESC~\citep{li2024escape}, ASC~\citep{aggarwal-etal-2023-lets}, and DSC~\citep{wang-etal-2025-make}. CGES uses five confidence signals (LNS~\citep{malinin2021uncertainty}, DeepConf~\citep{fu2026deep}, MARS~\citep{bakman-etal-2024-mars}) plus one near-ideal process-reward reference (CGES-RM, \texttt{Qwen2.5-Math-PRM-7B}~\citep{zhang2025lessonsdevelopingprocessreward}). All numbers are means over 10 seeds at $B{=}16$. For CGES rows in Table~\ref{tab:main_results}, we report the \emph{efficient} threshold: the smallest $\gamma$ within 0.2pp of SC accuracy. Pareto curves sweep $\gamma\in[0.7,0.9999]$. Full baseline descriptions, confidence definitions, hyperparameters, smaller-budget runs ($B{=}4,8$), and the $\gamma$ grid appear in Appendix~\ref{appendix:hyperparams}--\ref{appex:gamma_grid}.

\begin{table*}[t]
\centering
\scriptsize
\setlength{\tabcolsep}{3pt}
\caption{Main results at budget $B=16$. CGES rows use the efficient $\gamma$, the smallest $\gamma$ matching SC$\pm$0.2\,pp, or the largest $\gamma$ if no $\gamma$ reaches it. Mean$\pm$std across 10 seeds. $\Delta$ rows show difference vs.\ SC (\textcolor{ForestGreen}{green}=improvement, \textcolor{BrickRed}{red}=regression). Last column: avg over tasks. \textbf{Bold = best}, \underline{underline = second-best} per column.}
\label{tab:main_results}
\newcolumntype{C}{>{\centering\arraybackslash}p{11mm}}
\resizebox{\linewidth}{!}{%
\begin{tabular}{l|CC|CC|CC|CC|CC|CC}
\toprule
Method & \multicolumn{2}{c|}{AIME24} & \multicolumn{2}{c|}{GPQA} & \multicolumn{2}{c|}{MATH500} & \multicolumn{2}{c|}{GSM8K} & \multicolumn{2}{c|}{MMLU\_Pro} & \multicolumn{2}{c}{\textbf{Avg}} \\
\cmidrule(lr){2-3} \cmidrule(lr){4-5} \cmidrule(lr){6-7} \cmidrule(lr){8-9} \cmidrule(lr){10-11} \cmidrule(lr){12-13}
& Acc (\%) & \#Calls & Acc (\%) & \#Calls & Acc (\%) & \#Calls & Acc (\%) & \#Calls & Acc (\%) & \#Calls & Acc (\%) & \#Calls \\
\midrule
SC & 77.3$_{\pm1.3}$ & 16.00$_{\pm0.00}$ & \textbf{50.0$_{\pm0.6}$} & 16.00$_{\pm0.00}$ & 81.2$_{\pm0.8}$ & 16.00$_{\pm0.00}$ & 94.2$_{\pm0.2}$ & 16.00$_{\pm0.00}$ & \textbf{62.1$_{\pm0.3}$} & 16.00$_{\pm0.00}$ & \textbf{73.0} & 16.00 \\
ESC ($w$=4) & 77.3$_{\pm1.3}$ & 10.68$_{\pm0.41}$ & 48.0$_{\pm1.0}$ & 12.88$_{\pm0.19}$ & 81.1$_{\pm0.7}$ & 7.74$_{\pm0.07}$ & \underline{94.2$_{\pm0.2}$} & 5.16$_{\pm0.03}$ & 62.0$_{\pm0.4}$ & 8.99$_{\pm0.07}$ & 72.5 & 9.09 \\
\quad\small$\Delta$ & $0.0$ & \textcolor{ForestGreen}{$-5.3$} & \textcolor{BrickRed}{$-2.0$} & \textcolor{ForestGreen}{$-3.1$} & \textcolor{BrickRed}{$-0.1$} & \textcolor{ForestGreen}{$-8.3$} & $0.0$ & \textcolor{ForestGreen}{$-10.8$} & \textcolor{BrickRed}{$-0.1$} & \textcolor{ForestGreen}{$-7.0$} & \textcolor{BrickRed}{$-0.4$} & \textcolor{ForestGreen}{$-6.9$} \\
ESC ($w$=8) & 77.3$_{\pm1.3}$ & 13.73$_{\pm0.27}$ & 48.0$_{\pm1.0}$ & 15.20$_{\pm0.09}$ & 81.2$_{\pm0.8}$ & 11.42$_{\pm0.09}$ & 94.2$_{\pm0.2}$ & 9.41$_{\pm0.04}$ & \underline{62.1$_{\pm0.3}$} & 12.87$_{\pm0.04}$ & 72.6 & 12.53 \\
\quad\small$\Delta$ & $0.0$ & \textcolor{ForestGreen}{$-2.3$} & \textcolor{BrickRed}{$-2.0$} & \textcolor{ForestGreen}{$-0.8$} & $0.0$ & \textcolor{ForestGreen}{$-4.6$} & $0.0$ & \textcolor{ForestGreen}{$-6.6$} & $0.0$ & \textcolor{ForestGreen}{$-3.1$} & \textcolor{BrickRed}{$-0.4$} & \textcolor{ForestGreen}{$-3.5$} \\
ASC-$\beta$ & 77.3$_{\pm1.3}$ & 9.84$_{\pm0.34}$ & \underline{48.1$_{\pm1.0}$} & 12.37$_{\pm0.14}$ & 81.2$_{\pm0.8}$ & 7.40$_{\pm0.07}$ & 94.2$_{\pm0.2}$ & 5.00$_{\pm0.03}$ & 62.0$_{\pm0.3}$ & 8.55$_{\pm0.06}$ & 72.6 & 8.63 \\
\quad\small$\Delta$ & $0.0$ & \textcolor{ForestGreen}{$-6.2$} & \textcolor{BrickRed}{$-1.9$} & \textcolor{ForestGreen}{$-3.6$} & $0.0$ & \textcolor{ForestGreen}{$-8.6$} & $0.0$ & \textcolor{ForestGreen}{$-11.0$} & \textcolor{BrickRed}{$-0.1$} & \textcolor{ForestGreen}{$-7.4$} & \textcolor{BrickRed}{$-0.4$} & \textcolor{ForestGreen}{$-7.4$} \\
ASC-Dir & 77.3$_{\pm1.3}$ & 10.88$_{\pm0.36}$ & 48.1$_{\pm1.0}$ & 13.15$_{\pm0.25}$ & 81.3$_{\pm0.8}$ & 7.79$_{\pm0.13}$ & 94.2$_{\pm0.2}$ & 5.17$_{\pm0.04}$ & 62.0$_{\pm0.3}$ & 9.07$_{\pm0.06}$ & \underline{72.6} & 9.21 \\
\quad\small$\Delta$ & $0.0$ & \textcolor{ForestGreen}{$-5.1$} & \textcolor{BrickRed}{$-1.9$} & \textcolor{ForestGreen}{$-2.8$} & \textcolor{ForestGreen}{$+0.1$} & \textcolor{ForestGreen}{$-8.2$} & $0.0$ & \textcolor{ForestGreen}{$-10.8$} & \textcolor{BrickRed}{$-0.1$} & \textcolor{ForestGreen}{$-6.9$} & \textcolor{BrickRed}{$-0.4$} & \textcolor{ForestGreen}{$-6.8$} \\
DSC & 74.3$_{\pm2.1}$ & 6.43$_{\pm0.63}$ & 46.4$_{\pm2.5}$ & \textbf{6.69$_{\pm0.29}$} & 79.7$_{\pm0.4}$ & \underline{4.85$_{\pm0.06}$} & 93.9$_{\pm0.3}$ & 2.85$_{\pm0.06}$ & 60.9$_{\pm0.4}$ & \textbf{5.18$_{\pm0.10}$} & 71.0 & \textbf{5.20}$^{\dagger}$ \\
\quad\small$\Delta$ & \textcolor{BrickRed}{$-3.0$} & \textcolor{ForestGreen}{$-9.6$} & \textcolor{BrickRed}{$-3.6$} & \textcolor{ForestGreen}{$-9.3$} & \textcolor{BrickRed}{$-1.5$} & \textcolor{ForestGreen}{$-11.2$} & \textcolor{BrickRed}{$-0.3$} & \textcolor{ForestGreen}{$-13.2$} & \textcolor{BrickRed}{$-1.2$} & \textcolor{ForestGreen}{$-10.8$} & \textcolor{BrickRed}{$-1.9$} & \textcolor{ForestGreen}{$-10.8$} \\
\midrule
CGES-LNS (arith) & \textbf{77.7$_{\pm1.5}$} & 7.77$_{\pm0.32}$ & 47.8$_{\pm1.3}$ & 11.90$_{\pm0.16}$ & \underline{81.3$_{\pm0.5}$} & 5.99$_{\pm0.11}$ & 94.1$_{\pm0.3}$ & \underline{2.41$_{\pm0.02}$} & 61.9$_{\pm0.3}$ & 8.24$_{\pm0.05}$ & 72.6 & 7.26 \\
\quad\small$\Delta$ & \textcolor{ForestGreen}{$+0.4$} & \textcolor{ForestGreen}{$-8.2$} & \textcolor{BrickRed}{$-2.2$} & \textcolor{ForestGreen}{$-4.1$} & \textcolor{ForestGreen}{$+0.1$} & \textcolor{ForestGreen}{$-10.0$} & \textcolor{BrickRed}{$-0.1$} & \textcolor{ForestGreen}{$-13.6$} & \textcolor{BrickRed}{$-0.2$} & \textcolor{ForestGreen}{$-7.8$} & \textcolor{BrickRed}{$-0.4$} & \textcolor{ForestGreen}{$-8.7$} \\
CGES-LNS (geom) & 77.3$_{\pm1.3}$ & 6.05$_{\pm0.37}$ & 47.4$_{\pm1.1}$ & 13.62$_{\pm0.13}$ & 81.1$_{\pm0.3}$ & 5.65$_{\pm0.14}$ & 94.1$_{\pm0.2}$ & 3.04$_{\pm0.02}$ & 61.9$_{\pm0.4}$ & 8.95$_{\pm0.05}$ & 72.4 & 7.46 \\
\quad\small$\Delta$ & $0.0$ & \textcolor{ForestGreen}{$-9.9$} & \textcolor{BrickRed}{$-2.6$} & \textcolor{ForestGreen}{$-2.4$} & \textcolor{BrickRed}{$-0.1$} & \textcolor{ForestGreen}{$-10.3$} & \textcolor{BrickRed}{$-0.1$} & \textcolor{ForestGreen}{$-13.0$} & \textcolor{BrickRed}{$-0.2$} & \textcolor{ForestGreen}{$-7.1$} & \textcolor{BrickRed}{$-0.6$} & \textcolor{ForestGreen}{$-8.5$} \\
CGES-DeepConf (B10) & \underline{77.7$_{\pm1.5}$} & \underline{5.62$_{\pm0.37}$} & 47.9$_{\pm1.1}$ & 12.36$_{\pm0.15}$ & 81.0$_{\pm0.5}$ & 4.89$_{\pm0.10}$ & 94.1$_{\pm0.3}$ & 2.41$_{\pm0.02}$ & 61.9$_{\pm0.3}$ & 8.26$_{\pm0.05}$ & 72.5 & \underline{6.71} \\
\quad\small$\Delta$ & \textcolor{ForestGreen}{$+0.4$} & \textcolor{ForestGreen}{$-10.4$} & \textcolor{BrickRed}{$-2.1$} & \textcolor{ForestGreen}{$-3.6$} & \textcolor{BrickRed}{$-0.2$} & \textcolor{ForestGreen}{$-11.1$} & \textcolor{BrickRed}{$-0.1$} & \textcolor{ForestGreen}{$-13.6$} & \textcolor{BrickRed}{$-0.2$} & \textcolor{ForestGreen}{$-7.7$} & \textcolor{BrickRed}{$-0.4$} & \textcolor{ForestGreen}{$-9.3$} \\
CGES-DeepConf (tail) & 77.0$_{\pm1.8}$ & 8.64$_{\pm0.25}$ & 47.7$_{\pm1.2}$ & \underline{11.48$_{\pm0.15}$} & 81.2$_{\pm0.5}$ & 5.98$_{\pm0.14}$ & 94.1$_{\pm0.3}$ & 2.41$_{\pm0.02}$ & 61.9$_{\pm0.3}$ & \underline{8.23$_{\pm0.05}$} & 72.4 & 7.35 \\
\quad\small$\Delta$ & \textcolor{BrickRed}{$-0.3$} & \textcolor{ForestGreen}{$-7.4$} & \textcolor{BrickRed}{$-2.3$} & \textcolor{ForestGreen}{$-4.5$} & $0.0$ & \textcolor{ForestGreen}{$-10.0$} & \textcolor{BrickRed}{$-0.1$} & \textcolor{ForestGreen}{$-13.6$} & \textcolor{BrickRed}{$-0.2$} & \textcolor{ForestGreen}{$-7.8$} & \textcolor{BrickRed}{$-0.6$} & \textcolor{ForestGreen}{$-8.7$} \\
CGES-MARS & 77.3$_{\pm1.3}$ & \textbf{5.20$_{\pm0.34}$} & 47.9$_{\pm1.1}$ & 14.21$_{\pm0.31}$ & 81.3$_{\pm0.5}$ & 5.89$_{\pm0.13}$ & 94.2$_{\pm0.1}$ & 3.11$_{\pm0.03}$ & 62.0$_{\pm0.3}$ & 8.80$_{\pm0.05}$ & 72.6 & 7.44 \\
\quad\small$\Delta$ & $0.0$ & \textcolor{ForestGreen}{$-10.8$} & \textcolor{BrickRed}{$-2.1$} & \textcolor{ForestGreen}{$-1.8$} & \textcolor{ForestGreen}{$+0.1$} & \textcolor{ForestGreen}{$-10.1$} & $0.0$ & \textcolor{ForestGreen}{$-12.9$} & \textcolor{BrickRed}{$-0.1$} & \textcolor{ForestGreen}{$-7.2$} & \textcolor{BrickRed}{$-0.4$} & \textcolor{ForestGreen}{$-8.6$} \\
CGES-RM & 66.7$_{\pm0.0}$ & 13.60$_{\pm0.00}$ & 45.2$_{\pm2.3}$ & 14.51$_{\pm0.10}$ & \textbf{81.6$_{\pm0.8}$} & \textbf{2.88$_{\pm0.07}$} & \textbf{94.9$_{\pm0.4}$} & \textbf{1.25$_{\pm0.03}$} & 58.3$_{\pm0.6}$ & 12.31$_{\pm0.03}$ & 69.3 & 8.91 \\
\quad\small$\Delta$ & \textcolor{BrickRed}{$-10.6$} & \textcolor{ForestGreen}{$-2.4$} & \textcolor{BrickRed}{$-4.8$} & \textcolor{ForestGreen}{$-1.5$} & \textcolor{ForestGreen}{$+0.4$} & \textcolor{ForestGreen}{$-13.1$} & \textcolor{ForestGreen}{$+0.7$} & \textcolor{ForestGreen}{$-14.8$} & \textcolor{BrickRed}{$-3.8$} & \textcolor{ForestGreen}{$-3.7$} & \textcolor{BrickRed}{$-3.6$} & \textcolor{ForestGreen}{$-7.1$} \\
\bottomrule
\end{tabular}
}% end resizebox
\par\smallskip
{\scriptsize $^{\dagger}$DSC requires an additional LLM-based difficulty ranking step using the same inference model (${\approx}0.6$ amortized calls/question; each call ranks $B{=}5$ questions jointly), excluded from \#Calls but included in wall-clock time.}
\end{table*}

\begin{table}[t]
\centering
\scriptsize
\setlength{\tabcolsep}{3pt}
\caption{Wall-clock time per question (s) at $B=16$, assuming LLM batch size 1.
  Wall $=$ avg\_calls $\times$ ($T_{\rm LLM}$ + $T_{\rm conf}$).}
\label{tab:wallclock}
\resizebox{\columnwidth}{!}{%
\begin{tabular}{lrrrrrr}
\toprule
Method & AIME24 & GPQA & MATH500 & GSM8K & MMLU-Pro & Avg \\
\midrule
SC & 136.01 & 88.25 & 4.24 & 1.18 & 2.45 & 46.43 \\
ESC ($w$=4) & 90.80 & 71.03 & 2.05 & 0.38 & 1.38 & 33.13 \\
ESC ($w$=8) & 116.74 & 83.85 & 3.03 & 0.70 & 1.97 & 41.26 \\
ASC-$\beta$ & 83.65 & 68.21 & 1.96 & 0.37 & 1.31 & 31.10 \\
ASC-Dir & 92.46 & 72.54 & 2.07 & 0.38 & 1.39 & 33.77 \\
DSC & 54.86 & \textbf{36.98} & 1.32 & 0.24 & \textbf{0.84} & \textbf{18.85} \\
\midrule
CGES-LNS (arith) & 66.31 & 65.83 & 1.60 & \underline{0.18} & 1.27 & 27.04 \\
CGES-LNS (geom) & 51.57 & 75.37 & 1.50 & 0.23 & 1.38 & 26.01 \\
CGES-DeepConf (B10) & \underline{47.90} & 68.38 & \underline{1.30} & 0.18 & \underline{1.27} & \underline{23.81} \\
CGES-DeepConf (tail) & 73.68 & \underline{63.51} & 1.59 & 0.18 & 1.27 & 28.05 \\
CGES-MARS & \textbf{47.43} & 89.70 & 1.76 & 0.77 & 1.67 & 28.27 \\
CGES-RM & 124.64 & 86.00 & \textbf{0.98} & \textbf{0.16} & 2.80 & 42.92 \\
\bottomrule
\end{tabular}%
}
\end{table}

\subsection{CGES matches SC with about half the calls}
\label{subsec:main_results}

Table~\ref{tab:main_results} reports accuracy and call counts at $B{=}16$. The five probability-based CGES variants cut average calls from 16.0 to between 6.7 and 7.5 while staying within 0.2pp of SC accuracy on every benchmark except GPQA. The most efficient deployable variant, CGES-DeepConf (B10), uses \textbf{6.71} calls (a \textbf{58\%} reduction) at $-0.4$pp accuracy. ESC and ASC give smaller savings and lose 2pp on GPQA. DSC posts the lowest call count overall but at $-1.9$pp on average, a gap that widens at smaller budgets (Appendix Tables~\ref{tab:results_B4} and~\ref{tab:results_B8}). CGES-RM acts as a ceiling. When its training distribution matches the task it improves accuracy by 0.4pp on MATH500 (2.88 calls) and 0.7pp on GSM8K (1.25 calls), but accuracy drops outside that range, an effect we explain in Section~\ref{subsec:calibration}.

\paragraph{Call counts translate to real time.} SC is the only method with a fixed sample budget, so its $B$ samples can be drawn in one batched call. All adaptive methods, including ESC, ASC, DSC, and CGES, are sequential because each sampling decision depends on prior outputs. Thus, the relevant cost is sequential rounds rather than raw calls. Table~\ref{tab:wallclock} reports wall-clock time per question at batch size 1, where CGES still reduces latency by 40--50\% over SC, confirming that fewer calls translate into real time savings.

\begin{figure*}[t]
  \centering
  \includegraphics[width=\linewidth]{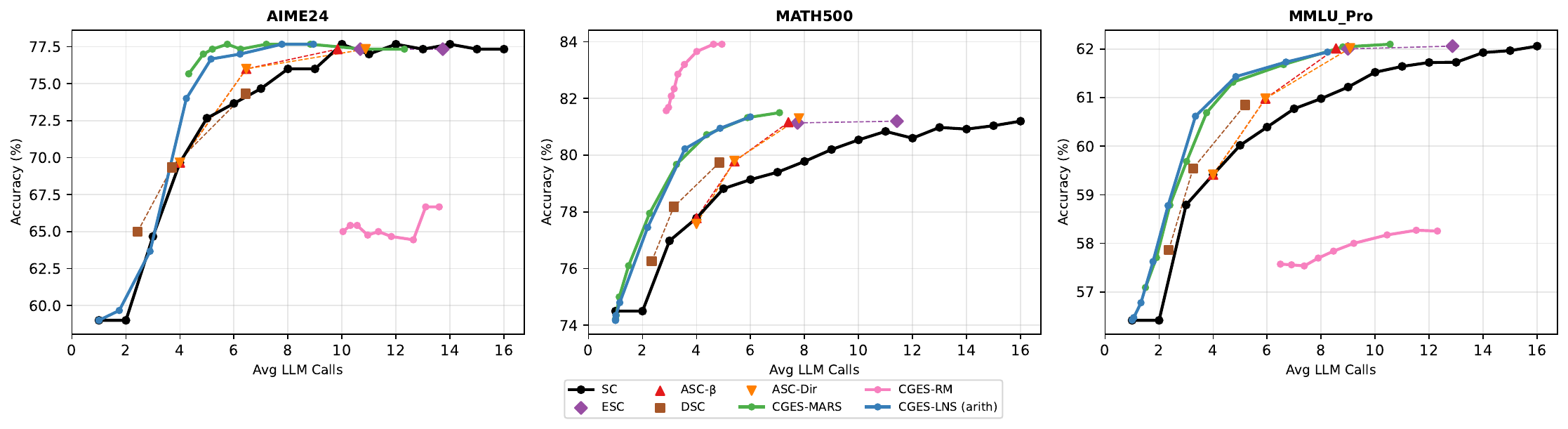}
  \caption{Accuracy vs.\ average LLM calls at $B{=}16$. SC (black) staircases through $B{=}4,8,16$. CGES-LNS / CGES-MARS (solid) sweep $\gamma\in[0.7,0.9999]$. DSC / ASC / ESC (dashed) plot one point per budget. Points above-left of SC are Pareto-dominant.}
  \label{fig:pareto_main}
\end{figure*}

\subsection{Pareto frontier}
\label{subsec:pareto}

Figure~\ref{fig:pareto_main} addresses matched-budget comparisons directly. The CGES curves Pareto-dominate SC and all stopping baselines across the full range on AIME24, MATH500, and MMLU\_Pro. CGES-RM on MATH500 sits well above every other curve, illustrating the ceiling. 

\subsection{When the Confidence Signal Is Uninformative}
\label{subsec:calibration}

Theorem~\ref{thm:realistic} shows CGES works only when the confidence signal is informative, that is, when it carries a positive drift toward correct answers. GPQA is the case where this fails. The confidence scores there barely separate correct from incorrect responses, so the drift is small, the posterior never concentrates, and CGES neither saves calls nor fully matches SC accuracy (Table~\ref{tab:main_results}). The other four benchmarks carry informative signals and CGES behaves as the theory predicts. This is not a weakness specific to one estimator. On GPQA every confidence signal is weakly informative, including the reward model. We measure signal informativeness per task and method, and confirm that it tracks CGES efficiency, in Appendix~\ref{appex:drift}.

\section{Conclusions}

We proposed CGES, a confidence-based Bayesian framework for test-time scaling of LLMs. By treating each response and its confidence as probabilistic evidence, CGES enables early stopping and more reliable aggregation than majority voting. Across five reasoning benchmarks, CGES substantially reduces model calls while maintaining or improving accuracy, outperforming Self-Consistency and early-stopping baselines. Our theoretical analysis further shows correctness under both ideal and noisy confidence assumptions. Overall, confidence-guided aggregation provides a principled solution within the family of test-time scaling methods that sample multiple responses and aggregate them into a single answer. Future work includes (i) developing more accurate confidence estimators to further enhance efficiency and accuracy, and (ii) predicting the required number of samples dynamically from confidence signals.

\FloatBarrier

\section*{Limitations}
CGES relies on confidence signals that may be imperfectly calibrated. While Theorem~\ref{thm:realistic} shows robustness to noisy estimates, strongly miscalibrated signals can degrade performance. Additionally, the framework assumes a finite candidate set $\mathcal{U}$; if the correct answer is absent from $\mathcal{U}$, no method operating solely on $\mathcal{U}$ can recover it. Future work includes developing more accurate confidence estimators and extending the framework with an explicit abstention hypothesis.

% \section*{Acknowledgments}
% Placeholder for acknowledgments.

% acl.sty already calls \bibliographystyle{acl_natbib} internally — do NOT repeat it here.
\bibliography{example_paper}

\appendix
\section{Full Proof of Lemma~\ref{lem:posterior_simplification}}
\label{appendix:lemma_proof}

\begin{proof}
By Bayes' rule,
\begin{equation}
\label{eq:bayes_start}
\mathbb{P}(I{=}i \mid \mathrm{Obs})
=
\frac{\mathbb{P}(\mathrm{Obs}\mid I{=}i)\,\mathbb{P}(I{=}i)}
{\sum_{k=1}^K \mathbb{P}(\mathrm{Obs}\mid I{=}k)\,\mathbb{P}(I{=}k)}.
\end{equation}
Under Assumption~\ref{assump:independence}, the likelihood factorizes as $\mathbb{P}(\mathrm{Obs}\mid I=i) = \prod_{t=1}^m \mathbb{P}(R_t,C_t\mid I=i)$.
Under the uniform prior (Assumption~\ref{assump:uniform}), $\mathbb{P}(I=i)=1/K$ is constant in $i$ and cancels between numerator and denominator of~\eqref{eq:bayes_start}, yielding
\begin{equation}
\label{eq:posterior_product_joint}
\mathbb{P}(I{=}i \mid \mathrm{Obs})
=
\frac{\prod_{t} \mathbb{P}(R_t,C_t\mid I{=}i)}
{\sum_{k} \prod_{t} \mathbb{P}(R_t,C_t\mid I{=}k)}.
\end{equation}
Next, by the chain rule, $\mathbb{P}(R_t,C_t\mid I=i)=\mathbb{P}(R_t\mid C_t,I=i)\,\mathbb{P}(C_t\mid I=i)$.
Assumption~\ref{assump:conf} implies $\mathbb{P}(C_t\mid I=i)$ is the same for all $i$, hence we may write $\mathbb{P}(C_t\mid I=i)=\mathbb{P}(C_t)$.
Substituting into~\eqref{eq:posterior_product_joint} gives
\[
\mathbb{P}(I{=}i \mid \mathrm{Obs})
\;=\;
\frac{\prod_{t} \mathbb{P}(R_t\mid C_t,I{=}i)\,\mathbb{P}(C_t)}
{\sum_{k} \prod_{t} \mathbb{P}(R_t\mid C_t,I{=}k)\,\mathbb{P}(C_t)}.
\]
The factor $\prod_{t=1}^m \mathbb{P}(C_t)$ is common to every term in numerator and denominator and cancels, yielding~\eqref{eq:pos}.
\end{proof}

\section{Full Proof of Theorem~\ref{thrm:ideal}}
\label{appendix:ideal_full_proof}

\begin{proof}
Without loss of generality, relabel so that the true index is $I=1$. Define
\begin{align*}
A_j &:= \prod_{t=1}^{m}\mathbb{P}(R_t\mid C_t, I=j),\\
X_j &:= A_j \,/\, {\textstyle\sum_{k=1}^{K}A_k}, \quad j\in[K].
\end{align*}
It suffices to show $A_1/A_k \xrightarrow{\text{a.s.}} \infty$ for every $k \neq 1$,
which implies $X_1 \xrightarrow{\text{a.s.}} 1$ and $X_k \xrightarrow{\text{a.s.}} 0$ for all $k\neq 1$.

\paragraph{Step 1.}
For a fixed $k\neq 1$, define
\begin{multline*}
Y_k^{(t)} := \log \mathbb{P}(R_t\mid C_t, I{=}1)\\
           - \log \mathbb{P}(R_t\mid C_t, I{=}k).
\end{multline*}
Under Assumption~\ref{assump:ideal}, conditioning on $C_t$,
\begin{align*}
\log \mathbb{P}(R_t\mid C_t, I=1)
&=
\begin{cases}
\log C_t & \text{if } R_t=a_1,\\
\log\!\big(\tfrac{1-C_t}{K-1}\big) & \text{if } R_t\neq a_1,
\end{cases}
\\[4pt]
\log \mathbb{P}(R_t\mid C_t, I{=}k)
&=
\begin{cases}
\log C_t & \text{if } R_t=a_k,\\
\log\!\big(\tfrac{1-C_t}{K-1}\big) & \text{if } R_t\neq a_k.
\end{cases}
\end{align*}
Let $\theta_t := \tfrac{1-C_t}{K-1}$. Under the true model $I=1$:
\begin{align*}
\mathbb{P}(R_t=a_1\mid C_t,I=1) &= C_t,\\
\mathbb{P}(R_t=a_k\mid C_t,I=1) &= \theta_t.
\end{align*}

\paragraph{Step 2.}
Taking the conditional expectation,
\[
\mathbb{E}\!\left[Y_k^{(t)}\mid C_t\right]
= (C_t-\theta_t)\log\!\Big(\tfrac{C_t}{\theta_t}\Big).
\]
Since $(a-b)\log(a/b)>0$ for $a\neq b$, this is strictly positive whenever $C_t\neq 1/K$. Noting $|Y_k^{(t)}| \le |\log(C_t/\theta_t)|$, the integrability assumption of Theorem~\ref{thrm:ideal} gives $\mathbb{E}|Y_k^{(t)}| < \infty$; combined with $\mathbb{P}(C_t = 1/K) = 0$,
\begin{align*}
\mu_k
&:= \mathbb{E}\!\left[Y_k^{(t)}\right]
 = \mathbb{E}\!\left[\mathbb{E}\!\left[Y_k^{(t)}\mid C_t\right]\right] \\
&> 0.
\end{align*}

\paragraph{Step 3.}
By Assumption~\ref{assump:independence}, $\{Y_k^{(t)}\}_{t=1}^m$ are i.i.d.\ with finite mean $\mu_k>0$.
The Strong Law of Large Numbers gives
$\frac{1}{m}\sum_{t=1}^{m} Y_k^{(t)} \xrightarrow{\text{a.s.}} \mu_k > 0$,
so $\sum_{t=1}^{m} Y_k^{(t)} \xrightarrow{\text{a.s.}} +\infty$.
Since $\log(A_1/A_k) = \sum_{t=1}^{m}Y_k^{(t)}$,
\[
\frac{A_1}{A_k} = \exp\!\Big(\textstyle\sum_{t=1}^{m}Y_k^{(t)}\Big) \xrightarrow{\text{a.s.}} \infty.
\]
Therefore $A_k/A_1\to 0$ a.s.\ for all $k\neq 1$, and so
\begin{align*}
X_1 &= \frac{1}{1+\sum_{k\neq 1} A_k/A_1} \xrightarrow{\text{a.s.}} 1,\\
X_k &= \frac{A_k/A_1}{1+\sum_{j\neq 1} A_j/A_1} \xrightarrow{\text{a.s.}} 0.
\end{align*}
\end{proof}

\section{Full Proof of Theorem~\ref{thm:realistic}}
\label{appendix:real_full_proof}

\begin{proof}
Without loss of generality, let $I=1$. Define
\begin{align*}
A_j &:= \prod_{t=1}^{m} \mathbb{P}(R_t\mid C_t, I=j),\\
X_j &:= A_j \,/\, {\textstyle\sum_{k=1}^{K}A_k}.
\end{align*}
It suffices to show $A_1/A_k \xrightarrow{\text{a.s.}} \infty$ for every $k\neq 1$,
implying $X_1\to 1$ and $X_k\to 0$ almost surely.

\paragraph{Step 1 (per-sample increment).}
Fix $k\neq 1$ and define
\begin{multline*}
Y_k^{(t)} := \log \mathbb{P}(R_t\mid C_t, I{=}1)\\
           - \log \mathbb{P}(R_t\mid C_t, I{=}k).
\end{multline*}
Writing $\theta_t=\tfrac{1-C_t}{K-1}$ and $\ell(C_t) := \log C_t - \log \theta_t = \log\!\tfrac{C_t(K-1)}{1-C_t}$, the ideal scoring rule of Eq.~\eqref{eq:ideal_conditional} gives
\[
Y_k^{(t)}
=
\bigl(\mathbf{1}\{R_t = a_1\} - \mathbf{1}\{R_t = a_k\}\bigr)\,\ell(C_t),
\]
since the contributions from indices $R_t \notin \{a_1, a_k\}$ cancel.

\paragraph{Step 2 (expectation under the true joint).}
Taking expectations under the true joint distribution of $(R_t, C_t)$,
\begin{align*}
\mathbb{E}\!\left[Y_k^{(t)}\right]
&= \mathbb{E}\!\left[\bigl(\mathbf{1}\{R_t {=} a_1\} - \mathbf{1}\{R_t {=} a_k\}\bigr)\ell(C_t)\right] \\
&= \mu_k.
\end{align*}
The integrability assumption $\mathbb{E}[|\ell(C_t)|]<\infty$ together with $|\mathbf{1}\{R_t = a_1\} - \mathbf{1}\{R_t = a_k\}| \le 1$ ensures $\mathbb{E}[|Y_k^{(t)}|]<\infty$. No further structural assumption on the joint distribution of $(R_t,C_t)$ is required.

\paragraph{Step 3 (SLLN).}
By Assumption~\ref{assump:independence}, $\{Y_k^{(t)}\}_{t=1}^m$ are i.i.d.\ with mean $\mu_k$. The Strong Law of Large Numbers gives
$\frac{1}{m}\sum_{t} Y_k^{(t)} \xrightarrow{\text{a.s.}} \mu_k$.
If $\mu_k > 0$, then $\sum_t Y_k^{(t)} \xrightarrow{\text{a.s.}} +\infty$ and hence
$A_1/A_k = \exp\!\bigl(\sum_t Y_k^{(t)}\bigr) \xrightarrow{\text{a.s.}} \infty$.
Combining over all $k \neq 1$, $X_1 \xrightarrow{\text{a.s.}} 1$ and $X_k \xrightarrow{\text{a.s.}} 0$.

\paragraph{Converse.}
If $\mu_{k^\star}<0$ for some $k^\star\neq 1$, then
$\sum_t Y_{k^\star}^{(t)} \xrightarrow{\text{a.s.}} -\infty$, so $A_1/A_{k^\star}\xrightarrow{\text{a.s.}} 0$ and $X_1\xrightarrow{\text{a.s.}} 0$.
\end{proof}

\section{Consistency of Dynamic-Support CGES}
\label{app:dynamic-support}

Theorems~\ref{thrm:ideal} and~\ref{thm:realistic} are stated for a fixed candidate set $\mathcal{U}$ and a fixed support size $K$. The online implementation of Algorithm~\ref{alg:cges} instead uses a growing candidate set: after $m$ samples it scores over the distinct answers observed so far together with the explicit OTHER/unseen bucket of Section~\ref{notations}. This section shows that the realistic-regime guarantee of Theorem~\ref{thm:realistic} extends to this dynamic-support version once the observed support stabilizes.

\paragraph{Setup.}
Let $R_t$ denote the concrete post-processed answer string produced at sample $t$, let $C_t\in(0,1)$ be its confidence score, and let $a^\star$ denote the true answer $A$. After $m$ samples, let $\mathcal{V}_m:=\{R_1,\ldots,R_m\}$ be the observed concrete support, $\mathcal{U}_m:=\mathcal{V}_m\cup\{\bot\}$ the dynamic candidate set, and $K_m:=|\mathcal{U}_m|$ its size, where $\bot$ denotes the OTHER/unseen category. Each observed sample is a concrete answer, so $R_t\neq\bot$ for all $t$. Consistent with the dynamic-support implementation described in Section~\ref{subsec:algo-score}, at each time $m$ the score is recomputed over all $m$ samples using the current candidate set $\mathcal{U}_m$ and current support size $K_m$. Writing $\theta_{m,t}:=(1-C_t)/(K_m-1)$ for the per-sample residual mass, the unnormalized and normalized scores of a candidate $a\in\mathcal{U}_m$ are
\begin{align*}
A_m(a) &:= \prod_{t=1}^m C_t^{\,\mathbf{1}\{R_t=a\}}\,\theta_{m,t}^{\,\mathbf{1}\{R_t\neq a\}},\\
X_m(a) &:= \frac{A_m(a)}{\sum_{b\in\mathcal{U}_m}A_m(b)}.
\end{align*}
For a support size $K$, write $\ell_K(c):=\log\frac{c(K-1)}{1-c}$; this is the per-sample log-likelihood ratio $\ell$ of Theorem~\ref{thm:realistic} with the support size made explicit.

\begin{proposition}[Dynamic-support consistency after support stabilization]
\label{prop:dynamic-support-stabilization}
Fix a realization of the observation sequence $\{(R_t,C_t)\}_{t\ge 1}$, and suppose its observed candidate set stabilizes: there exist a finite time $T<\infty$ and a finite candidate set $\mathcal{U}_\infty$ such that $\mathcal{U}_m=\mathcal{U}_\infty$ for all $m\ge T$. Let $K_\infty:=|\mathcal{U}_\infty|$, suppose the true answer satisfies $a^\star\in\mathcal{U}_\infty$, and for each competitor $b\in\mathcal{U}_\infty\setminus\{a^\star\}$ define the per-sample log-ratio increment
\begin{multline*}
    Y_t^{(b)}
    := \bigl(\mathbf{1}\{R_t=a^\star\}-\mathbf{1}\{R_t=b\}\bigr)\\
    \cdot\,\ell_{K_\infty}(C_t).
\end{multline*}
If, for every such $b$,
\[
    \liminf_{m\to\infty}\;\frac{1}{m}\sum_{t=1}^m Y_t^{(b)}\;>\;0,
\]
then $X_m(a^\star)\to 1$ and $X_m(b)\to 0$ for every $b\in\mathcal{U}_\infty\setminus\{a^\star\}$.
\end{proposition}

\begin{proof}
For all $m\ge T$ the candidate set has stabilized: $\mathcal{U}_m=\mathcal{U}_\infty$ and $K_m=K_\infty$. Fix any competitor $b\in\mathcal{U}_\infty\setminus\{a^\star\}$. For $m\ge T$ every factor in $A_m(\cdot)$ uses $K_\infty$, so the log-score ratio between $a^\star$ and $b$ is
\[
    \log\frac{A_m(a^\star)}{A_m(b)}=\sum_{t=1}^m Y_t^{(b)}.
\]
To see this, consider one sample $t$ and write $\theta_{\infty,t}:=(1-C_t)/(K_\infty-1)$. If $R_t=a^\star$, the contribution to $A_m(a^\star)$ is $C_t$ and to $A_m(b)$ is $\theta_{\infty,t}$, so the log-ratio contribution is $\ell_{K_\infty}(C_t)$. If $R_t=b$, the same computation gives $-\ell_{K_\infty}(C_t)$. If $R_t\notin\{a^\star,b\}$, both candidates receive the identical factor $\theta_{\infty,t}$, so the contribution is zero. This also covers $b=\bot$, since $R_t\neq\bot$ for every observed sample. By the assumed positive empirical drift, $\sum_{t=1}^m Y_t^{(b)}\to+\infty$, hence $A_m(b)/A_m(a^\star)\to 0$ for every $b\in\mathcal{U}_\infty\setminus\{a^\star\}$. Since $\mathcal{U}_\infty$ is finite and $X_m(a^\star)=\bigl(1+\sum_{b\neq a^\star}A_m(b)/A_m(a^\star)\bigr)^{-1}$, it follows that $X_m(a^\star)\to 1$, and hence $X_m(b)\to 0$ for every $b\neq a^\star$.
\end{proof}

Proposition~\ref{prop:dynamic-support-stabilization} is a deterministic statement about a single observation sequence. The next corollary gives a stochastic condition under which its hypotheses, and hence consistency, hold almost surely.

\begin{corollary}[Finite-support dynamic CGES]
\label{cor:finite-support-dynamic-cges}
Assume $\{(R_t,C_t)\}_{t\ge 1}$ are i.i.d.\ (Assumption~\ref{assump:independence}) and the concrete answer distribution has finite support $\mathcal{V}:=\{a:\mathbb{P}(R_t=a)>0\}$, with $a^\star\in\mathcal{V}$. Then $\mathcal{U}_m$ stabilizes almost surely to $\mathcal{U}_\infty=\mathcal{V}\cup\{\bot\}$. Let $K_\infty:=|\mathcal{V}|+1$ and assume $\mathbb{E}\bigl[|\ell_{K_\infty}(C_t)|\bigr]<\infty$. Let $Y_t^{(b)}$ be the increment of Proposition~\ref{prop:dynamic-support-stabilization}, and for the OTHER bucket set $Y_t^{(\bot)}:=\mathbf{1}\{R_t=a^\star\}\,\ell_{K_\infty}(C_t)$, which is the $b=\bot$ instance since $R_t\neq\bot$. If the per-sample drifts are positive, that is, $\mu_b:=\mathbb{E}[Y_t^{(b)}]>0$ for every concrete competitor $b\in\mathcal{V}\setminus\{a^\star\}$ and $\mu_\bot:=\mathbb{E}[Y_t^{(\bot)}]>0$, then $X_m(a^\star)\xrightarrow{\mathrm{a.s.}}1$ and $X_m(b)\xrightarrow{\mathrm{a.s.}}0$ for every $b\in\mathcal{U}_\infty\setminus\{a^\star\}$.
\end{corollary}

\begin{proof}
For any fixed $a\in\mathcal{V}$, let $p_a:=\mathbb{P}(R_t=a)>0$. Then $\mathbb{P}(a\notin\mathcal{V}_m)=(1-p_a)^m\to 0$, so the first hitting time $T_a:=\inf\{m\ge 1:R_m=a\}$ is finite almost surely. Since $\mathcal{V}$ is finite, $T:=\max_{a\in\mathcal{V}}T_a$ is finite almost surely, and $\mathcal{U}_m=\mathcal{V}\cup\{\bot\}=\mathcal{U}_\infty$ for all $m\ge T$. The stabilization hypothesis of Proposition~\ref{prop:dynamic-support-stabilization} therefore holds almost surely.

It remains to verify the positive empirical drift. Fix a concrete competitor $b\in\mathcal{V}\setminus\{a^\star\}$. The integrability assumption gives $\mathbb{E}|Y_t^{(b)}|\le\mathbb{E}[|\ell_{K_\infty}(C_t)|]<\infty$, and by hypothesis $\mathbb{E}[Y_t^{(b)}]=\mu_b>0$. Since the samples are i.i.d., the Strong Law of Large Numbers gives $\frac{1}{m}\sum_{t=1}^m Y_t^{(b)}\xrightarrow{\mathrm{a.s.}}\mu_b>0$. The same argument applied to $Y_t^{(\bot)}$ gives $\frac{1}{m}\sum_{t=1}^m Y_t^{(\bot)}\xrightarrow{\mathrm{a.s.}}\mu_\bot>0$. Thus the positive-drift hypothesis of Proposition~\ref{prop:dynamic-support-stabilization} holds almost surely for every competitor in $\mathcal{U}_\infty\setminus\{a^\star\}$, and the claim follows.
\end{proof}

\begin{corollary}[Eventual threshold crossing]
\label{cor:dynamic-support-threshold-crossing}
Under the hypotheses of Proposition~\ref{prop:dynamic-support-stabilization}, fix any threshold $\gamma\in(0,1)$. There exists a finite time $M_\gamma$ such that $X_m(a^\star)\ge\gamma$ and $a^\star\in\arg\max_{a\in\mathcal{U}_m}X_m(a)$ for all $m\ge M_\gamma$. Consequently, a dynamic-support CGES run with no finite budget, or with finite budget $B\ge M_\gamma$, that has not stopped before $M_\gamma$, stops no later than $M_\gamma$ and returns $a^\star$. If $B<M_\gamma$, the run may instead stop at budget exhaustion before the threshold is crossed.
\end{corollary}

\begin{proof}
By Proposition~\ref{prop:dynamic-support-stabilization}, $X_m(a^\star)\to 1$ and $X_m(b)\to 0$ for every $b\in\mathcal{U}_\infty\setminus\{a^\star\}$. Since $\gamma<1$, there is a finite $M_1$ with $X_m(a^\star)\ge\gamma$ for all $m\ge M_1$. Since $\mathcal{U}_\infty$ is finite and every competing score converges to zero, there is a finite $M_2$ with $X_m(a^\star)>X_m(b)$ for every $b\in\mathcal{U}_\infty\setminus\{a^\star\}$ and all $m\ge M_2$. Set $M_\gamma:=\max\{M_1,M_2,T\}$. For all $m\ge M_\gamma$ the support has stabilized, $a^\star$ meets the threshold, and $a^\star$ is the strict argmax. The stopping claim then follows for a run with no finite budget or with $B\ge M_\gamma$; if $B<M_\gamma$, the run may stop at budget exhaustion before $m$ reaches $M_\gamma$.
\end{proof}

\begin{remark}[Role of the OTHER bucket]
The condition $\mu_\bot>0$ is needed when $\bot$ is included in the normalized score and in the stopping rule $\max_{a\in\mathcal{U}_m}X_m(a)\ge\gamma$. If $\bot$ is used only as a bookkeeping device and the prediction is restricted to concrete observed answers, consistency of the concrete argmax requires only $\mu_b>0$ for every $b\in\mathcal{V}\setminus\{a^\star\}$.
\end{remark}

\begin{remark}[Scope of the result]
In the CGES setting the stabilization hypothesis of Proposition~\ref{prop:dynamic-support-stabilization} holds automatically. An LLM with a finite vocabulary and a bounded generation length emits answers from a finite set, so the observed candidate set $\mathcal{U}_m$ is a non-decreasing family of subsets of a finite universe and necessarily stabilizes after a finite, run-dependent time on every realization. Consequently the dynamic-support extension introduces no assumption beyond those used for the fixed-support Theorem~\ref{thm:realistic}; in particular, the finite-support hypothesis of Corollary~\ref{cor:finite-support-dynamic-cges} is always met.

The genuine finite-budget caveat is different. Proposition~\ref{prop:dynamic-support-stabilization} is asymptotic and does not rule out premature stopping before $a^\star$ has entered the observed support: if $a^\star\notin\mathcal{U}_m$, no rule restricted to $\mathcal{U}_m$ can return it. This is intrinsic to any method whose candidate set is built from observed samples; in finite-budget runs the risk is controlled through the stopping threshold and can be reduced further by requiring a small minimum number of samples before early stopping is allowed.
\end{remark}

\section{Full Proof of Proposition~\ref{prop:risk_bound}}
\label{appendix:risk_bound_proof}

\begin{proof}
Let $\mathcal{F}_m := \sigma\!\bigl(\{(R_t, C_t)\}_{t=1}^m\bigr)$ denote the natural filtration generated by the first $m$ samples, and write $\mathrm{Obs}_m = \{(R_t, C_t)\}_{t=1}^m$. We write $X_i^{(m)}$ for the score on candidate $a_i$ computed by Algorithm~\ref{alg:cges} from $\mathrm{Obs}_m$.

\paragraph{Step 1 (Posterior identification).}
By Lemma~\ref{lem:posterior_simplification}, Assumptions~\ref{assump:independence}, \ref{assump:uniform}, and~\ref{assump:conf} imply that the true Bayesian posterior over the correct index $I$ given $\mathrm{Obs}_m$ admits the closed form of Eq.~\eqref{eq:pos}. Under Assumption~\ref{assump:ideal}, the conditional likelihood $\mathbb{P}(R_t \mid C_t, I = i)$ takes the explicit one-vs-rest form of Eq.~\eqref{eq:ideal_conditional}, which is precisely the auxiliary likelihood used by Algorithm~\ref{alg:cges}. Consequently, the normalized score satisfies
\begin{equation}
\label{eq:posterior_identity}
X_i^{(m)} \;=\; \mathbb{P}(I = i \mid \mathrm{Obs}_m),
\end{equation}
for every $i \in [K]$ and every $m \ge 1$.

\paragraph{Step 2 (Stopping time).}
Define $\tau_\gamma := \inf\{m \ge 1 : \max_{i \in [K]} X_i^{(m)} \ge \gamma\}$, with $\inf \emptyset := +\infty$. Since $\max_i X_i^{(m)}$ is $\mathcal{F}_m$-measurable for each $m$, the event $\{\tau_\gamma \le m\} = \bigcup_{j = 1}^m \{\max_i X_i^{(j)} \ge \gamma\}$ lies in $\mathcal{F}_m$, so $\tau_\gamma$ is a stopping time with respect to $\{\mathcal{F}_m\}_{m \ge 1}$. Algorithm~\ref{alg:cges} terminates at $T := \min(\tau_\gamma, B)$ and returns
\[
\hat{I} \;:=\; \arg\max_{i \in [K]} X_i^{(T)},
\]
breaking ties arbitrarily among maximizers.

\paragraph{Step 3 (Posterior identity at the stopping time).}
Since $T$ is bounded by $B$, the events $\{T = m\}$ for $m = 1, \ldots, B$ partition the sample space; each is $\mathcal{F}_m$-measurable because $T$ is a stopping time. Combining this partition with the deterministic-time identity~\eqref{eq:posterior_identity},
\begin{equation}
\label{eq:posterior_at_stopping}
\begin{aligned}
\mathbb{P}(I {=} i \mid \mathcal{F}_T)
&= \sum_{m=1}^{B} \mathbf{1}\{T{=}m\}\,\mathbb{P}(I {=} i \mid \mathcal{F}_m) \\
&= \sum_{m=1}^{B} \mathbf{1}\{T{=}m\}\,X_i^{(m)} \\
&= X_i^{(T)},
\end{aligned}
\end{equation}
so the posterior identity extends from deterministic times to the random stopping time $T$.

\paragraph{Step 4 (Risk bound on the threshold-stopping event).}
On the event $\{\tau_\gamma \le B\}$, we have $T = \tau_\gamma$ and, by the definition of $\tau_\gamma$, $\max_i X_i^{(\tau_\gamma)} \ge \gamma$. Combining~\eqref{eq:posterior_at_stopping} with the definition of $\hat{I}$,
\begin{align*}
\mathbb{P}\!\bigl(I {=} \hat{I} \,\bigm|\, \mathcal{F}_{\tau_\gamma}\bigr)
&= X_{\hat{I}}^{(\tau_\gamma)} \\
&= \max_{i \in [K]} X_i^{(\tau_\gamma)}
 \;\ge\; \gamma.
\end{align*}
Taking the complement yields
\begin{align*}
\mathbb{P}\!\bigl(I \ne \hat{I} \,\bigm|\, \mathcal{F}_{\tau_\gamma}\bigr)
&= 1 - X_{\hat{I}}^{(\tau_\gamma)} \\
&\le 1 - \gamma,
\end{align*}
which is the bound in the proposition.
\end{proof}

\paragraph{Remark (budget-exhaustion event).}
The bound in Proposition~\ref{prop:risk_bound} applies on $\{\tau_\gamma \le B\}$, where the algorithm halts because the posterior on the top candidate exceeds $\gamma$. On the complementary event $\{\tau_\gamma > B\}$, Algorithm~\ref{alg:cges} halts at the budget with $\max_i X_i^{(B)} < \gamma$; the exact identity $\mathbb{P}(I \ne \hat{I} \mid \mathrm{Obs}_B) = 1 - \max_i X_i^{(B)}$ continues to hold by Step~1 (which is independent of the stopping rule), but the conditional risk is no longer controlled by $1 - \gamma$. Empirically, the budget-exhaustion event is rare for moderate $\gamma$ and $B$ (Section~\ref{experiments}).

\begin{corollary}[Bayes-optimal decision and earliest certified stopping]
\label{cor:optimality}
Under Assumptions~\ref{assump:independence}--\ref{assump:conf}, for every finite $m \ge 1$ and any (possibly randomized) decision rule $\delta_m$ measurable with respect to $\mathcal{F}_m$,
\[
\mathbb{P}\!\bigl(\delta_m \ne I \,\bigm|\, \mathcal{F}_m\bigr) \;\ge\; 1 - \max_{i \in [K]} X_i^{(m)},
\]
with equality attained by $\hat{I}_m \in \arg\max_i X_i^{(m)}$. The same inequality holds at any bounded $\{\mathcal{F}_m\}$-stopping time $T \le B$, with $\mathcal{F}_m$ replaced by $\mathcal{F}_T$. Consequently, for $\gamma \in (0,1)$ and $\tau_\gamma := \inf\{m \ge 1 : \max_i X_i^{(m)} \ge \gamma\}$, any $\{\mathcal{F}_m\}$-measurable rule whose conditional success probability at its bounded stopping time $T$ is at least $\gamma$ must satisfy $T \ge \tau_\gamma$ almost surely.
\end{corollary}

\begin{proof}
By Step~1 of the proof of Proposition~\ref{prop:risk_bound}, $X_i^{(m)} = \mathbb{P}(I = i \mid \mathcal{F}_m)$. Conditional on $\mathcal{F}_m$, the randomization in $\delta_m$ is independent of $I$, so
\begin{align*}
\mathbb{P}\!\bigl(\delta_m = I \,\bigm|\, \mathcal{F}_m\bigr)
&= \sum_{i=1}^{K} \mathbb{P}(\delta_m = i \mid \mathcal{F}_m)\, X_i^{(m)} \\
&\le \max_{i \in [K]} X_i^{(m)},
\end{align*}
with equality whenever $\delta_m$ is supported on $\arg\max_i X_i^{(m)}$, in particular for $\hat{I}_m$. Taking the complement gives the displayed inequality at deterministic times. Step~3 of the same proof established the stopped-time identity $\mathbb{P}(I = i \mid \mathcal{F}_T) = X_i^{(T)}$ for every bounded $\{\mathcal{F}_m\}$-stopping time $T \le B$; the same conditional argument applied at $\mathcal{F}_T$ yields the inequality at $T$. Finally, $\mathbb{P}(\delta_T = I \mid \mathcal{F}_T) \ge \gamma$ forces $\max_i X_i^{(T)} \ge \gamma$, and by the definition of $\tau_\gamma$ this can occur only when $T \ge \tau_\gamma$.
\end{proof}

\FloatBarrier

\section{Hyperparameters and Decoding}
\label{appendix:hyperparams}

For all benchmarks we use temperature 0.7 and top-$p{=}0.95$ for sampling. Self-consistency runs with budget $B{\in}\{4,8,16\}$. ESC uses window sizes $w{\in}\{4,8\}$. ASC uses the original authors' threshold settings. For DSC we use the difficulty bucket schedule from~\citet{wang-etal-2025-make}. Each table cell reports the mean and standard deviation over 10 random seeds.

\begin{figure}[h]
    \centering
    \begin{subfigure}[t]{0.48\textwidth}
        \centering
        \includegraphics[width=\linewidth]{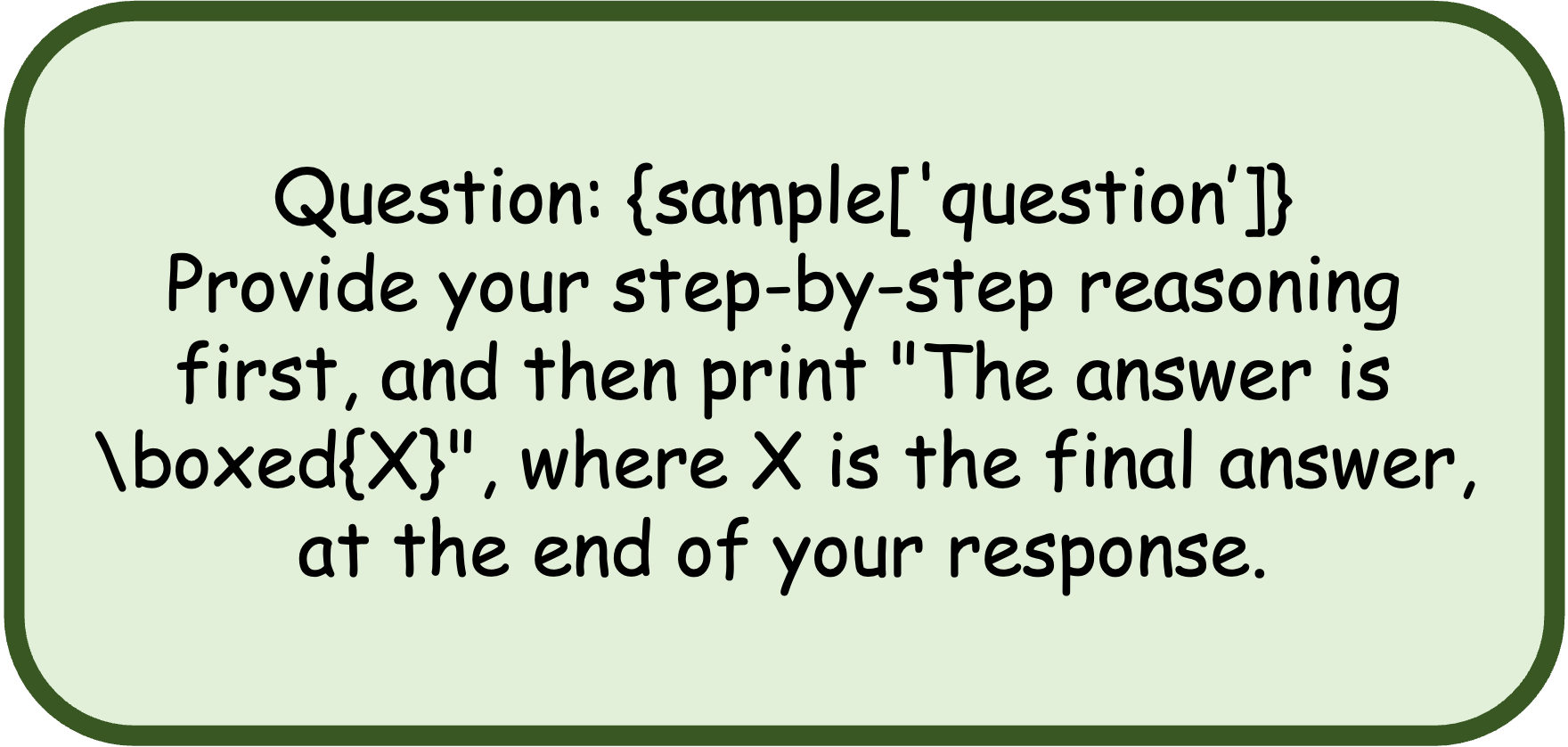}
        \subcaption{Prompt used for \textbf{MATH500, AIME24, GSM8K}}
    \end{subfigure}
    \hfill
    \begin{subfigure}[t]{0.48\textwidth}
        \centering
        \includegraphics[width=\linewidth]{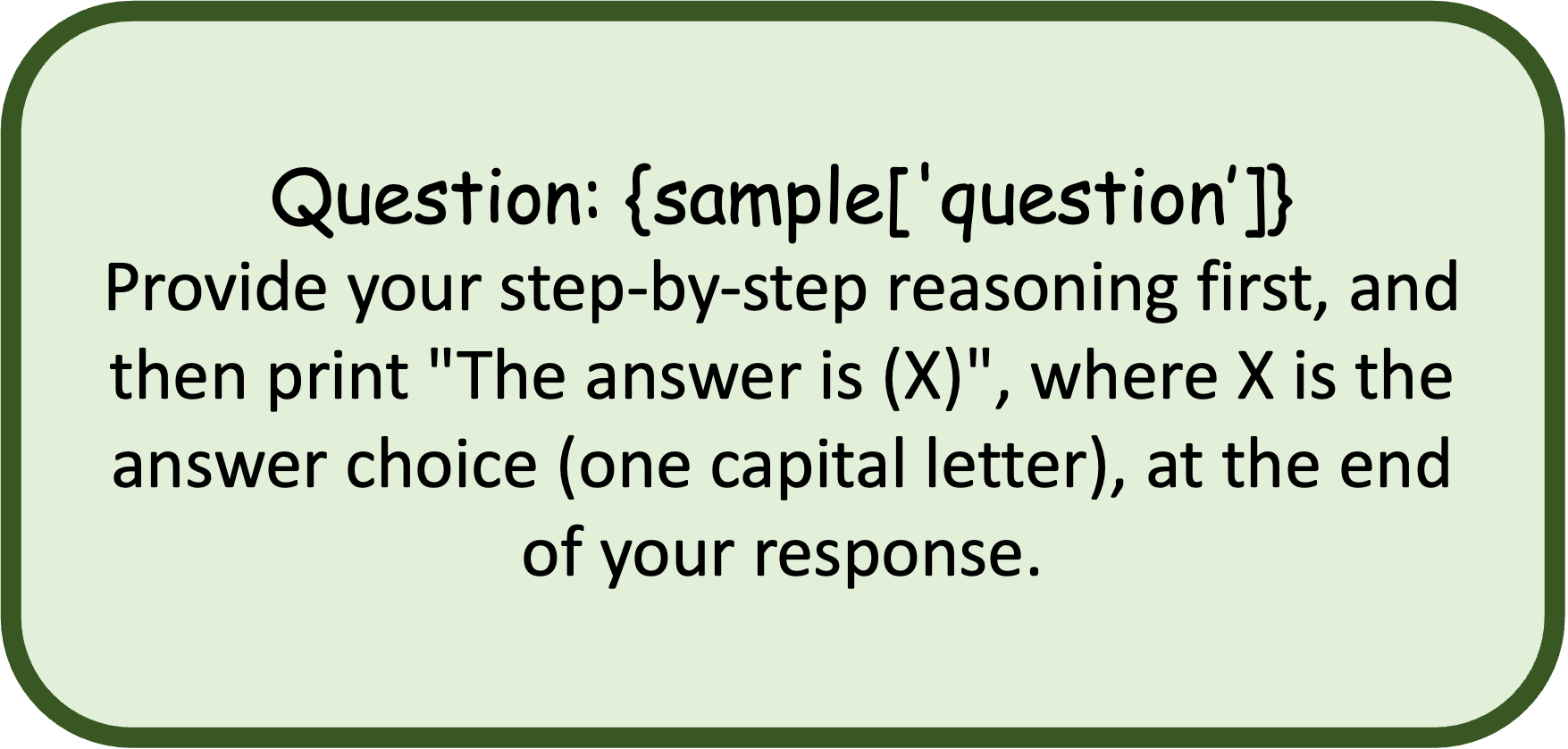}
        \subcaption{Prompt used for \textbf{GPQA, MMLU\_Pro}}
    \end{subfigure}
    \caption{Two prompt templates for evaluation.}
    \label{fig:prompts}
\end{figure}

\paragraph{Prompt Templates.} Dataset-specific answer-format constraints simplify parsing. Prompt templates are shown in Figure~\ref{fig:prompts}.

\section{Baselines}
\label{appex:baselines}

\textbf{Self-Consistency (SC)}~\citep{wang2023selfconsistency} samples a fixed budget of $B$ responses and returns the majority vote. \textbf{Early-Stopping Self-Consistency (ESC)}~\citep{li2024escape} halts once the last $w$ predictions agree. \textbf{Adaptive-Consistency (ASC)}~\citep{aggarwal-etal-2023-lets} stops once a Bayesian estimate of the majority is stable. We use the Beta-binomial variant (ASC-$\beta$) and the Dirichlet-multinomial variant (ASC-Dir). \textbf{Difficulty-Adaptive Self-Consistency (DSC)}~\citep{wang-etal-2025-make} adds an LLM-based difficulty ranking step that allocates more calls to harder questions. The ranking step is excluded from \#Calls but included in the wall-clock numbers in Table~\ref{tab:wallclock}.

\section{Confidence Estimation Strategies}
\label{subsec:confidence-estimation}

We compare several strategies for estimating the \emph{scalar} confidence $C_t \in (0,1)$ of each sampled answer $R_t$. Each strategy maps model outputs (token probabilities or reward scores) to a scalar $C_t$ that we interpret as an estimate of the probability that the corresponding response $R_t$ is correct. These estimates need not be perfectly calibrated. They serve as noisy confidence signals in the sense formalized in Section~\ref{notations} and Theorem~\ref{thm:realistic}. Our framework operates at the response level, but confidence estimation relies on finer token-level granularities. Specifically, let a response $R_t$ consist of a token sequence $\{T_1,\dots,T_L\}$ with associated autoregressive probabilities $\{p_1,\dots,p_L\}$. We use lowercase $p_\ell$ to denote these token-level probabilities from the underlying language model, and uppercase $\mathbb{P}(\cdot)$ for probabilities over candidate answers $a_j \in \mathcal{U}$ as in Section~\ref{notations}. The role of the present section is to map token-level scores (or reward-model outputs) into a single scalar confidence $C_t$. Building on this, we describe three token-based approaches and one verifier-based alternative.

\paragraph{Length-Normalized Scoring (LNS)~\citep{malinin2021uncertainty}.}
A natural way to quantify the likelihood of a response is by averaging over token probabilities. The \emph{geometric mean} yields the standard length-normalized score $\text{LNS}_{\text{geom}} = \exp\!\big(\tfrac{1}{L}\sum_{\ell=1}^L \log p_\ell\big)$, while the \emph{arithmetic mean} provides a simpler length-insensitive proxy, $\text{LNS}_{\text{arith}} = \tfrac{1}{L}\sum_{\ell=1}^L p_\ell$. We set $C_t$ to either of these values and denote them in results as \emph{LNS (geom)} and \emph{LNS (arith)}.

\paragraph{DeepConf~\citep{fu2026deep}.}
DeepConf forms a confidence signal from the least certain parts of the generation rather than the whole sequence, which makes it sensitive to local reasoning errors. We use two variants. \emph{DeepConf-B10} averages the lowest-probability tokens, taking the mean over the bottom 10\% of $\{p_1,\dots,p_L\}$. \emph{DeepConf-Tail} averages the token probabilities in the final segment of the response (the last 20\% of tokens), where the answer is committed. Both are squashed to $(0,1)$ and set as $C_t$, denoted \emph{DeepConf-B10} and \emph{DeepConf-Tail}.

\paragraph{MARS (Step-Weighted Scoring)~\citep{bakman-etal-2024-mars}.}
The MARS method generalizes LNS by assigning different weights to different positions in the sequence. Each token $T_\ell$ receives an exponent $w(R, Q, L, \ell) \triangleq \frac{1}{2L} + \frac{u(R, Q, \ell)}{2}$, where $R$ denotes the full textual response string, so the overall score becomes $\bar{P}(R \mid Q, \theta) = \prod_{\ell=1}^L p_\ell^{\,w(R, Q, L, \ell)}$, where $\theta$ denotes the parameters of the language model. Here $u(R, Q, \ell)$ is an importance score for token $T_\ell$, such as the semantic change in the output when masking that token. To address the expense of per-token weighting for long reasoning responses, we adopt a \emph{step-wise} variant of MARS. Instead of per-token weights, we assign weights at the granularity of reasoning steps or sentence segments. Although the step-importance score is recomputed each iteration, this overhead (6 layers, $\sim$50M parameters) is negligible compared to the 7B-parameter inference model we query for $R_t$. We denote the resulting confidence as $C_t = \text{MARS}$.

\paragraph{Reward Model Confidence.}
In addition to the above token-level methods, we also consider a model-based approach that evaluates entire responses directly. A trained reward model assigns a quality score to each $R_t$, which we use as $C_t$. In our study we use the \texttt{Qwen2.5-Math-PRM-7B} process reward model~\citep{zhang2025lessonsdevelopingprocessreward}, which outputs a scalar in $(0,1)$ that correlates with alignment to the ground truth on math-style reasoning. We include it as a near-optimal reference rather than a deployable setting, since the PRM is trained specifically for step-level scoring and its training distribution overlaps with several benchmarks, so its signal approximates an upper bound on confidence quality on in-domain tasks. We denote this variant as \emph{RM Confidence}.

\section{Additional Results at Smaller Budgets}
\label{appex:add_res}

We re-run the main experiments at $B{=}4$ and $B{=}8$ (Tables~\ref{tab:results_B4} and~\ref{tab:results_B8}). ESC is omitted at $B{=}4$ because its minimum window size $w{\geq}4$ exceeds the budget. The picture is the same as at $B{=}16$. CGES variants match SC accuracy within a small margin while using fewer calls. CGES-RM on math tasks improves accuracy further. DSC stays the most aggressive on call count but the accuracy gap to SC widens to $-2.7$pp at $B{=}4$ and $-3.1$pp at $B{=}8$.

\begin{table*}[t]
\centering
\scriptsize
\setlength{\tabcolsep}{3pt}
\caption{Results at budget $B=4$. CGES rows use the efficient $\gamma$ (smallest $\gamma$ matching SC$\pm$0.2\,pp). Mean$\pm$std across 10 seeds. $\Delta$ rows show difference vs.\ SC (\textcolor{ForestGreen}{green}=improvement, \textcolor{BrickRed}{red}=regression). Last column: avg over tasks. Bold = best, underline = second-best per column. ESC omitted: its minimum window size ($w{\geq}4$) exceeds this budget, making comparison meaningless.}
\label{tab:results_B4}
\newcolumntype{C}{>{\centering\arraybackslash}p{11mm}}
\resizebox{\linewidth}{!}{%
\begin{tabular}{l|CC|CC|CC|CC|CC|CC}
\toprule
Method & \multicolumn{2}{c|}{AIME24} & \multicolumn{2}{c|}{GPQA} & \multicolumn{2}{c|}{MATH500} & \multicolumn{2}{c|}{GSM8K} & \multicolumn{2}{c|}{MMLU\_Pro} & \multicolumn{2}{c}{\textbf{Avg}} \\
\cmidrule(lr){2-3} \cmidrule(lr){4-5} \cmidrule(lr){6-7} \cmidrule(lr){8-9} \cmidrule(lr){10-11} \cmidrule(lr){12-13}
& Acc (\%) & \#Calls & Acc (\%) & \#Calls & Acc (\%) & \#Calls & Acc (\%) & \#Calls & Acc (\%) & \#Calls & Acc (\%) & \#Calls \\
\midrule
SC & 69.7$_{\pm5.5}$ & 4.00$_{\pm0.00}$ & \textbf{44.9$_{\pm1.8}$} & 4.00$_{\pm0.00}$ & 77.8$_{\pm0.9}$ & 4.00$_{\pm0.00}$ & 93.3$_{\pm0.4}$ & 4.00$_{\pm0.00}$ & 59.4$_{\pm0.6}$ & 4.00$_{\pm0.00}$ & 69.0 & 4.00 \\
ASC-$\beta$ & 69.7$_{\pm5.5}$ & 4.00$_{\pm0.00}$ & 42.5$_{\pm1.9}$ & 4.00$_{\pm0.00}$ & 77.8$_{\pm0.9}$ & 4.00$_{\pm0.00}$ & 93.3$_{\pm0.4}$ & 4.00$_{\pm0.00}$ & 59.4$_{\pm0.6}$ & 4.00$_{\pm0.00}$ & 68.5 & 4.00 \\
\quad\small$\Delta$ & $0.0$ & $0.0$ & \textcolor{BrickRed}{$-2.4$} & $0.0$ & $0.0$ & $0.0$ & $0.0$ & $0.0$ & $0.0$ & $0.0$ & \textcolor{BrickRed}{$-0.5$} & $0.0$ \\
ASC-Dir & 69.7$_{\pm5.5}$ & 4.00$_{\pm0.00}$ & 42.5$_{\pm1.9}$ & 4.00$_{\pm0.00}$ & 77.6$_{\pm0.8}$ & 4.00$_{\pm0.00}$ & 93.3$_{\pm0.4}$ & 4.00$_{\pm0.00}$ & 59.4$_{\pm0.6}$ & 4.00$_{\pm0.00}$ & 68.5 & 4.00 \\
\quad\small$\Delta$ & $0.0$ & $0.0$ & \textcolor{BrickRed}{$-2.4$} & $0.0$ & \textcolor{BrickRed}{$-0.2$} & $0.0$ & $0.0$ & $0.0$ & $0.0$ & $0.0$ & \textcolor{BrickRed}{$-0.5$} & $0.0$ \\
DSC & 65.0$_{\pm4.8}$ & \textbf{2.43$_{\pm0.09}$} & 39.8$_{\pm1.8}$ & \textbf{2.46$_{\pm0.04}$} & 76.3$_{\pm1.1}$ & \underline{2.34$_{\pm0.01}$} & 92.6$_{\pm0.3}$ & 2.10$_{\pm0.01}$ & 57.9$_{\pm0.9}$ & \textbf{2.36$_{\pm0.01}$} & 66.3 & \textbf{2.34}$^{\dagger}$ \\
\quad\small$\Delta$ & \textcolor{BrickRed}{$-4.7$} & \textcolor{ForestGreen}{$-1.6$} & \textcolor{BrickRed}{$-5.1$} & \textcolor{ForestGreen}{$-1.5$} & \textcolor{BrickRed}{$-1.5$} & \textcolor{ForestGreen}{$-1.7$} & \textcolor{BrickRed}{$-0.7$} & \textcolor{ForestGreen}{$-1.9$} & \textcolor{BrickRed}{$-1.5$} & \textcolor{ForestGreen}{$-1.6$} & \textcolor{BrickRed}{$-2.7$} & \textcolor{ForestGreen}{$-1.7$} \\
\midrule
CGES-LNS (arith) & 69.7$_{\pm3.8}$ & 3.53$_{\pm0.06}$ & \underline{44.2$_{\pm1.5}$} & 4.00$_{\pm0.00}$ & 78.2$_{\pm0.9}$ & 2.58$_{\pm0.02}$ & \underline{93.4$_{\pm0.4}$} & 1.83$_{\pm0.01}$ & 59.5$_{\pm0.7}$ & \underline{2.75$_{\pm0.02}$} & 69.0 & 2.94 \\
\quad\small$\Delta$ & $0.0$ & \textcolor{ForestGreen}{$-0.5$} & \textcolor{BrickRed}{$-0.7$} & $0.0$ & \textcolor{ForestGreen}{$+0.4$} & \textcolor{ForestGreen}{$-1.4$} & \textcolor{ForestGreen}{$+0.1$} & \textcolor{ForestGreen}{$-2.2$} & \textcolor{ForestGreen}{$+0.1$} & \textcolor{ForestGreen}{$-1.2$} & $0.0$ & \textcolor{ForestGreen}{$-1.1$} \\
CGES-LNS (geom) & \textbf{70.7$_{\pm4.2}$} & 2.89$_{\pm0.14}$ & 44.0$_{\pm1.8}$ & 4.00$_{\pm0.00}$ & 77.9$_{\pm0.9}$ & 2.38$_{\pm0.01}$ & 93.3$_{\pm0.4}$ & \underline{1.63$_{\pm0.02}$} & 59.3$_{\pm0.7}$ & 3.01$_{\pm0.01}$ & \underline{69.1} & 2.78 \\
\quad\small$\Delta$ & \textcolor{ForestGreen}{$+1.0$} & \textcolor{ForestGreen}{$-1.1$} & \textcolor{BrickRed}{$-0.9$} & $0.0$ & \textcolor{ForestGreen}{$+0.1$} & \textcolor{ForestGreen}{$-1.6$} & $0.0$ & \textcolor{ForestGreen}{$-2.4$} & \textcolor{BrickRed}{$-0.1$} & \textcolor{ForestGreen}{$-1.0$} & $0.0$ & \textcolor{ForestGreen}{$-1.2$} \\
CGES-DeepConf (B10) & 70.0$_{\pm4.5}$ & 2.97$_{\pm0.11}$ & 44.1$_{\pm2.0}$ & 4.00$_{\pm0.00}$ & \underline{78.5$_{\pm0.9}$} & 2.59$_{\pm0.02}$ & 93.4$_{\pm0.4}$ & 1.83$_{\pm0.01}$ & \underline{59.5$_{\pm0.6}$} & 2.77$_{\pm0.02}$ & \textbf{69.1} & 2.83 \\
\quad\small$\Delta$ & \textcolor{ForestGreen}{$+0.3$} & \textcolor{ForestGreen}{$-1.0$} & \textcolor{BrickRed}{$-0.8$} & $0.0$ & \textcolor{ForestGreen}{$+0.7$} & \textcolor{ForestGreen}{$-1.4$} & \textcolor{ForestGreen}{$+0.1$} & \textcolor{ForestGreen}{$-2.2$} & \textcolor{ForestGreen}{$+0.1$} & \textcolor{ForestGreen}{$-1.2$} & \textcolor{ForestGreen}{$+0.1$} & \textcolor{ForestGreen}{$-1.2$} \\
CGES-DeepConf (tail) & 69.3$_{\pm4.9}$ & 3.98$_{\pm0.03}$ & 43.5$_{\pm2.2}$ & 4.00$_{\pm0.01}$ & 78.1$_{\pm0.8}$ & 2.57$_{\pm0.02}$ & 93.4$_{\pm0.4}$ & 1.83$_{\pm0.01}$ & 59.5$_{\pm0.7}$ & 2.75$_{\pm0.02}$ & 68.8 & 3.03 \\
\quad\small$\Delta$ & \textcolor{BrickRed}{$-0.4$} & $0.0$ & \textcolor{BrickRed}{$-1.4$} & $0.0$ & \textcolor{ForestGreen}{$+0.3$} & \textcolor{ForestGreen}{$-1.4$} & \textcolor{ForestGreen}{$+0.1$} & \textcolor{ForestGreen}{$-2.2$} & \textcolor{ForestGreen}{$+0.1$} & \textcolor{ForestGreen}{$-1.2$} & \textcolor{BrickRed}{$-0.3$} & \textcolor{ForestGreen}{$-1.0$} \\
CGES-MARS & \underline{70.3$_{\pm3.8}$} & \underline{2.67$_{\pm0.09}$} & 43.4$_{\pm1.7}$ & 4.00$_{\pm0.00}$ & 77.8$_{\pm1.0}$ & 2.48$_{\pm0.01}$ & 93.4$_{\pm0.4}$ & 1.69$_{\pm0.01}$ & \textbf{59.5$_{\pm0.6}$} & 3.03$_{\pm0.01}$ & 68.9 & \underline{2.78} \\
\quad\small$\Delta$ & \textcolor{ForestGreen}{$+0.6$} & \textcolor{ForestGreen}{$-1.3$} & \textcolor{BrickRed}{$-1.5$} & $0.0$ & $0.0$ & \textcolor{ForestGreen}{$-1.5$} & \textcolor{ForestGreen}{$+0.1$} & \textcolor{ForestGreen}{$-2.3$} & \textcolor{ForestGreen}{$+0.1$} & \textcolor{ForestGreen}{$-1.0$} & \textcolor{BrickRed}{$-0.1$} & \textcolor{ForestGreen}{$-1.2$} \\
CGES-RM & 52.3$_{\pm3.4}$ & 3.98$_{\pm0.01}$ & 42.5$_{\pm2.0}$ & \underline{3.99$_{\pm0.01}$} & \textbf{80.2$_{\pm0.6}$} & \textbf{1.63$_{\pm0.02}$} & \textbf{94.4$_{\pm0.5}$} & \textbf{1.13$_{\pm0.01}$} & 59.2$_{\pm0.6}$ & 3.91$_{\pm0.00}$ & 65.7 & 2.93 \\
\quad\small$\Delta$ & \textcolor{BrickRed}{$-17.4$} & \textcolor{ForestGreen}{$-0.0$} & \textcolor{BrickRed}{$-2.4$} & \textcolor{ForestGreen}{$-0.0$} & \textcolor{ForestGreen}{$+2.4$} & \textcolor{ForestGreen}{$-2.4$} & \textcolor{ForestGreen}{$+1.1$} & \textcolor{ForestGreen}{$-2.9$} & \textcolor{BrickRed}{$-0.2$} & \textcolor{ForestGreen}{$-0.1$} & \textcolor{BrickRed}{$-3.3$} & \textcolor{ForestGreen}{$-1.1$} \\
\bottomrule
\end{tabular}
}% end resizebox
\par\smallskip
{\scriptsize $^{\dagger}$DSC requires an additional LLM-based difficulty ranking step using the same inference model (${\approx}0.6$ amortized calls/question; each call ranks $B{=}5$ questions jointly), excluded from \#Calls but included in wall-clock time.}
\end{table*}

\begin{table*}[t]
\centering
\scriptsize
\setlength{\tabcolsep}{3pt}
\caption{Results at budget $B=8$. CGES rows use the efficient $\gamma$ (smallest $\gamma$ matching SC$\pm$0.2\,pp). Mean$\pm$std across 10 seeds. $\Delta$ rows show difference vs.\ SC (\textcolor{ForestGreen}{green}=improvement, \textcolor{BrickRed}{red}=regression). Last column: avg over tasks. Bold = best, underline = second-best per column. ESC omitted: at $B=8$ its smallest window ($w{=}4$) leaves only two windows, too few for its early-stopping rule to be meaningful.}
\label{tab:results_B8}
\newcolumntype{C}{>{\centering\arraybackslash}p{11mm}}
\resizebox{\linewidth}{!}{%
\begin{tabular}{l|CC|CC|CC|CC|CC|CC}
\toprule
Method & \multicolumn{2}{c|}{AIME24} & \multicolumn{2}{c|}{GPQA} & \multicolumn{2}{c|}{MATH500} & \multicolumn{2}{c|}{GSM8K} & \multicolumn{2}{c|}{MMLU\_Pro} & \multicolumn{2}{c}{\textbf{Avg}} \\
\cmidrule(lr){2-3} \cmidrule(lr){4-5} \cmidrule(lr){6-7} \cmidrule(lr){8-9} \cmidrule(lr){10-11} \cmidrule(lr){12-13}
& Acc (\%) & \#Calls & Acc (\%) & \#Calls & Acc (\%) & \#Calls & Acc (\%) & \#Calls & Acc (\%) & \#Calls & Acc (\%) & \#Calls \\
\midrule
SC & 76.0$_{\pm2.9}$ & 8.00$_{\pm0.00}$ & \textbf{47.9$_{\pm2.1}$} & 8.00$_{\pm0.00}$ & 79.8$_{\pm0.6}$ & 8.00$_{\pm0.00}$ & 93.9$_{\pm0.2}$ & 8.00$_{\pm0.00}$ & 61.0$_{\pm0.4}$ & 8.00$_{\pm0.00}$ & \textbf{71.7} & 8.00 \\
ASC-$\beta$ & 76.0$_{\pm2.9}$ & 6.46$_{\pm0.10}$ & 46.0$_{\pm2.1}$ & \underline{7.22$_{\pm0.07}$} & 79.8$_{\pm0.6}$ & 5.41$_{\pm0.03}$ & 93.9$_{\pm0.2}$ & 4.49$_{\pm0.01}$ & 61.0$_{\pm0.4}$ & 5.94$_{\pm0.02}$ & 71.3 & 5.90 \\
\quad\small$\Delta$ & $0.0$ & \textcolor{ForestGreen}{$-1.5$} & \textcolor{BrickRed}{$-1.9$} & \textcolor{ForestGreen}{$-0.8$} & $0.0$ & \textcolor{ForestGreen}{$-2.6$} & $0.0$ & \textcolor{ForestGreen}{$-3.5$} & $0.0$ & \textcolor{ForestGreen}{$-2.1$} & \textcolor{BrickRed}{$-0.4$} & \textcolor{ForestGreen}{$-2.1$} \\
ASC-Dir & 76.0$_{\pm2.9}$ & 6.46$_{\pm0.10}$ & 46.0$_{\pm2.1}$ & 7.22$_{\pm0.07}$ & 79.8$_{\pm0.6}$ & 5.41$_{\pm0.03}$ & 93.9$_{\pm0.2}$ & 4.49$_{\pm0.01}$ & 61.0$_{\pm0.4}$ & 5.94$_{\pm0.02}$ & 71.3 & 5.90 \\
\quad\small$\Delta$ & $0.0$ & \textcolor{ForestGreen}{$-1.5$} & \textcolor{BrickRed}{$-1.9$} & \textcolor{ForestGreen}{$-0.8$} & $0.0$ & \textcolor{ForestGreen}{$-2.6$} & $0.0$ & \textcolor{ForestGreen}{$-3.5$} & $0.0$ & \textcolor{ForestGreen}{$-2.1$} & \textcolor{BrickRed}{$-0.4$} & \textcolor{ForestGreen}{$-2.1$} \\
DSC & 69.3$_{\pm4.7}$ & \textbf{3.71$_{\pm0.27}$} & 42.8$_{\pm2.6}$ & \textbf{3.76$_{\pm0.11}$} & 78.2$_{\pm0.7}$ & 3.16$_{\pm0.03}$ & 93.4$_{\pm0.3}$ & 2.34$_{\pm0.02}$ & 59.5$_{\pm0.5}$ & \textbf{3.26$_{\pm0.04}$} & 68.7 & \textbf{3.25}$^{\dagger}$ \\
\quad\small$\Delta$ & \textcolor{BrickRed}{$-6.7$} & \textcolor{ForestGreen}{$-4.3$} & \textcolor{BrickRed}{$-5.1$} & \textcolor{ForestGreen}{$-4.2$} & \textcolor{BrickRed}{$-1.6$} & \textcolor{ForestGreen}{$-4.8$} & \textcolor{BrickRed}{$-0.5$} & \textcolor{ForestGreen}{$-5.7$} & \textcolor{BrickRed}{$-1.5$} & \textcolor{ForestGreen}{$-4.7$} & \textcolor{BrickRed}{$-3.1$} & \textcolor{ForestGreen}{$-4.8$} \\
\midrule
CGES-LNS (arith) & 76.3$_{\pm2.3}$ & 4.29$_{\pm0.24}$ & 46.4$_{\pm1.8}$ & 7.67$_{\pm0.02}$ & 79.8$_{\pm0.4}$ & 3.17$_{\pm0.05}$ & \underline{94.0$_{\pm0.3}$} & 2.35$_{\pm0.02}$ & \underline{61.1$_{\pm0.4}$} & 4.37$_{\pm0.02}$ & 71.5 & \underline{4.37} \\
\quad\small$\Delta$ & \textcolor{ForestGreen}{$+0.3$} & \textcolor{ForestGreen}{$-3.7$} & \textcolor{BrickRed}{$-1.5$} & \textcolor{ForestGreen}{$-0.3$} & $0.0$ & \textcolor{ForestGreen}{$-4.8$} & \textcolor{ForestGreen}{$+0.1$} & \textcolor{ForestGreen}{$-5.7$} & \textcolor{ForestGreen}{$+0.1$} & \textcolor{ForestGreen}{$-3.6$} & \textcolor{BrickRed}{$-0.2$} & \textcolor{ForestGreen}{$-3.6$} \\
CGES-LNS (geom) & \underline{76.7$_{\pm2.6}$} & 4.43$_{\pm0.26}$ & 46.3$_{\pm1.6}$ & 7.98$_{\pm0.01}$ & \underline{80.1$_{\pm0.3}$} & 3.63$_{\pm0.05}$ & 93.9$_{\pm0.3}$ & \underline{2.22$_{\pm0.02}$} & 61.0$_{\pm0.5}$ & 4.46$_{\pm0.02}$ & 71.6 & 4.54 \\
\quad\small$\Delta$ & \textcolor{ForestGreen}{$+0.7$} & \textcolor{ForestGreen}{$-3.6$} & \textcolor{BrickRed}{$-1.6$} & $0.0$ & \textcolor{ForestGreen}{$+0.3$} & \textcolor{ForestGreen}{$-4.4$} & $0.0$ & \textcolor{ForestGreen}{$-5.8$} & $0.0$ & \textcolor{ForestGreen}{$-3.5$} & \textcolor{BrickRed}{$-0.1$} & \textcolor{ForestGreen}{$-3.5$} \\
CGES-DeepConf (B10) & \textbf{76.7$_{\pm1.5}$} & 4.61$_{\pm0.20}$ & \underline{46.6$_{\pm1.7}$} & 7.82$_{\pm0.01}$ & 79.8$_{\pm0.3}$ & 3.19$_{\pm0.04}$ & 94.0$_{\pm0.3}$ & 2.35$_{\pm0.02}$ & \textbf{61.1$_{\pm0.4}$} & 4.39$_{\pm0.02}$ & \underline{71.6} & 4.47 \\
\quad\small$\Delta$ & \textcolor{ForestGreen}{$+0.7$} & \textcolor{ForestGreen}{$-3.4$} & \textcolor{BrickRed}{$-1.3$} & \textcolor{ForestGreen}{$-0.2$} & $0.0$ & \textcolor{ForestGreen}{$-4.8$} & \textcolor{ForestGreen}{$+0.1$} & \textcolor{ForestGreen}{$-5.7$} & \textcolor{ForestGreen}{$+0.1$} & \textcolor{ForestGreen}{$-3.6$} & \textcolor{BrickRed}{$-0.1$} & \textcolor{ForestGreen}{$-3.5$} \\
CGES-DeepConf (tail) & 76.3$_{\pm3.1}$ & 4.88$_{\pm0.20}$ & 46.2$_{\pm1.9}$ & 7.56$_{\pm0.04}$ & 79.6$_{\pm0.2}$ & \underline{3.16$_{\pm0.04}$} & 94.0$_{\pm0.3}$ & 2.35$_{\pm0.02}$ & 61.0$_{\pm0.4}$ & \underline{4.37$_{\pm0.02}$} & 71.4 & 4.46 \\
\quad\small$\Delta$ & \textcolor{ForestGreen}{$+0.3$} & \textcolor{ForestGreen}{$-3.1$} & \textcolor{BrickRed}{$-1.7$} & \textcolor{ForestGreen}{$-0.4$} & \textcolor{BrickRed}{$-0.2$} & \textcolor{ForestGreen}{$-4.8$} & \textcolor{ForestGreen}{$+0.1$} & \textcolor{ForestGreen}{$-5.7$} & $0.0$ & \textcolor{ForestGreen}{$-3.6$} & \textcolor{BrickRed}{$-0.3$} & \textcolor{ForestGreen}{$-3.5$} \\
CGES-MARS & 76.3$_{\pm2.8}$ & \underline{4.12$_{\pm0.21}$} & 46.3$_{\pm1.4}$ & 8.00$_{\pm0.01}$ & 80.1$_{\pm0.4}$ & 3.81$_{\pm0.06}$ & 93.9$_{\pm0.3}$ & 2.24$_{\pm0.02}$ & 60.9$_{\pm0.4}$ & 4.38$_{\pm0.02}$ & 71.5 & 4.51 \\
\quad\small$\Delta$ & \textcolor{ForestGreen}{$+0.3$} & \textcolor{ForestGreen}{$-3.9$} & \textcolor{BrickRed}{$-1.6$} & $0.0$ & \textcolor{ForestGreen}{$+0.3$} & \textcolor{ForestGreen}{$-4.2$} & $0.0$ & \textcolor{ForestGreen}{$-5.8$} & \textcolor{BrickRed}{$-0.1$} & \textcolor{ForestGreen}{$-3.6$} & \textcolor{BrickRed}{$-0.2$} & \textcolor{ForestGreen}{$-3.5$} \\
CGES-RM & 52.7$_{\pm4.7}$ & 7.46$_{\pm0.06}$ & 43.7$_{\pm1.2}$ & 7.88$_{\pm0.03}$ & \textbf{80.8$_{\pm0.8}$} & \textbf{2.13$_{\pm0.04}$} & \textbf{94.8$_{\pm0.5}$} & \textbf{1.19$_{\pm0.02}$} & 58.5$_{\pm0.4}$ & 7.16$_{\pm0.02}$ & 66.1 & 5.16 \\
\quad\small$\Delta$ & \textcolor{BrickRed}{$-23.3$} & \textcolor{ForestGreen}{$-0.5$} & \textcolor{BrickRed}{$-4.2$} & \textcolor{ForestGreen}{$-0.1$} & \textcolor{ForestGreen}{$+1.0$} & \textcolor{ForestGreen}{$-5.9$} & \textcolor{ForestGreen}{$+0.9$} & \textcolor{ForestGreen}{$-6.8$} & \textcolor{BrickRed}{$-2.5$} & \textcolor{ForestGreen}{$-0.8$} & \textcolor{BrickRed}{$-5.6$} & \textcolor{ForestGreen}{$-2.8$} \\
\bottomrule
\end{tabular}
}% end resizebox
\par\smallskip
{\scriptsize $^{\dagger}$DSC requires an additional LLM-based difficulty ranking step using the same inference model (${\approx}0.6$ amortized calls/question; each call ranks $B{=}5$ questions jointly), excluded from \#Calls but included in wall-clock time.}
\end{table*}

\begin{figure}[t]
  \centering
  \includegraphics[width=\linewidth]{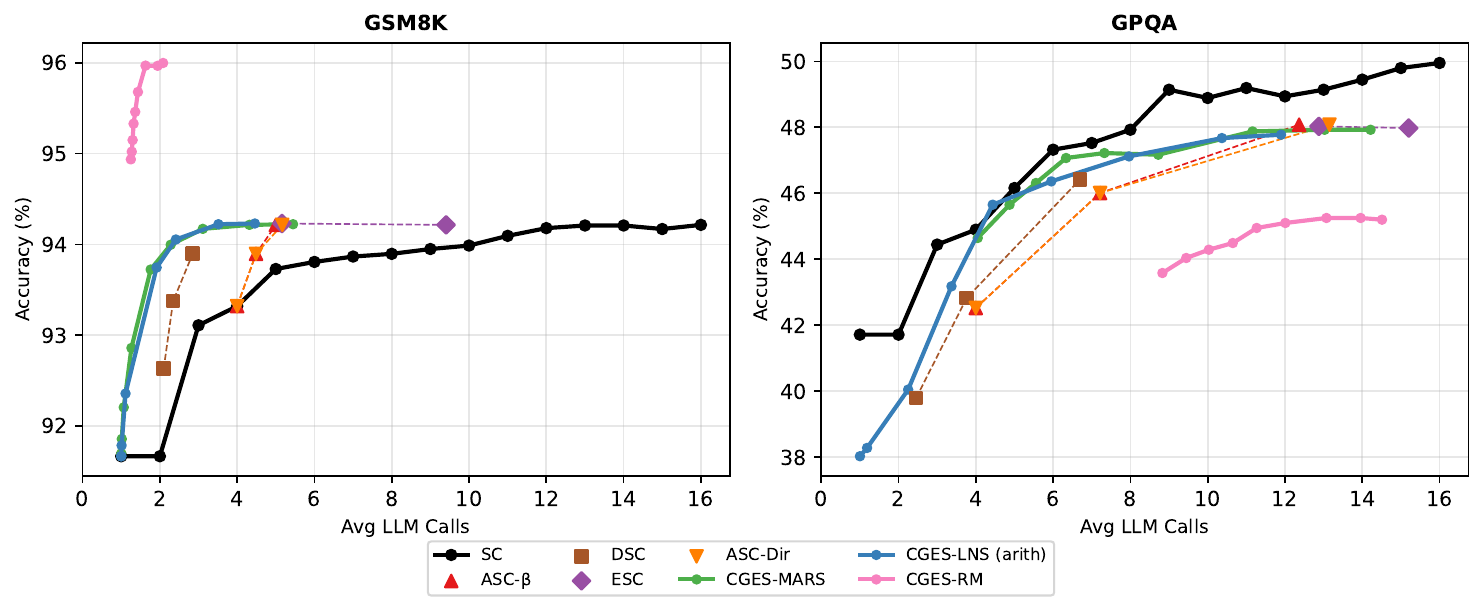}
  \caption{Pareto frontier for GSM8K and GPQA (same setup as Figure~\ref{fig:pareto_main}). CGES dominates the baselines on GSM8K where token-probability confidence is well-calibrated. On GPQA the available signals do not discriminate well, so CGES tracks SC without improving on it.}
  \label{fig:pareto_appendix}
\end{figure}

Figure~\ref{fig:pareto_appendix} adds GSM8K and GPQA. On GSM8K, CGES-RM reaches roughly 95\% accuracy at 1.2 calls, illustrating what a strong confidence signal enables. On GPQA, no variant strictly dominates SC, which matches the calibration analysis in Appendix~\ref{appex:drift}.

\section{Confidence Signal Quality and CGES Efficiency}
\label{appex:drift}

The main body claims that CGES helps only when the confidence signal is informative, and that GPQA is the case where it is not. This appendix makes that claim quantitative. We measure signal quality in three ways and show each one tracks CGES efficiency, defined as $1 - \overline{\#\text{Calls}}/B$, where $\overline{\#\text{Calls}}$ is the average number of LLM calls per question and $B$ is the budget.

\paragraph{Calibration and discrimination.} A confidence signal can be good in two distinct ways. \emph{Calibration}, measured by Expected Calibration Error (ECE), asks whether the average confidence matches the true accuracy. \emph{Discrimination}, measured by AUC-ROC, asks whether the signal ranks correct responses above incorrect ones. The drift $\mu_k$ in Theorem~\ref{thm:realistic} depends on the second, not the first. A signal can be well calibrated yet carry no drift if it cannot separate correct from incorrect responses. Table~\ref{tab:calib_auc} reports both quantities and CGES efficiency for every (method, task) pair.

\paragraph{GPQA is the uninformative case.} Table~\ref{tab:calib_auc} shows that efficiency follows discrimination, not calibration. The sharpest evidence is CGES-RM on GPQA versus MATH500. The two have the same ECE (0.050), so they look equally well calibrated, yet their AUC differs sharply (0.685 vs.\ 0.954) and so does efficiency (0.128 vs.\ 0.711). On GPQA the reward model assigns scores narrowly clustered around the overall accuracy regardless of correctness, which makes ECE small but leaves AUC near random. Every other signal on GPQA shows the same pattern of near-random AUC, so the drift stays small for all of them and CGES neither saves calls nor fully matches SC accuracy there, exactly as Theorem~\ref{thm:realistic} predicts for a weak drift. AIME24 is a second hard case for the reward model, where both ECE and AUC are poor.

\begin{figure}[t]
  \centering
  \includegraphics[width=\linewidth]{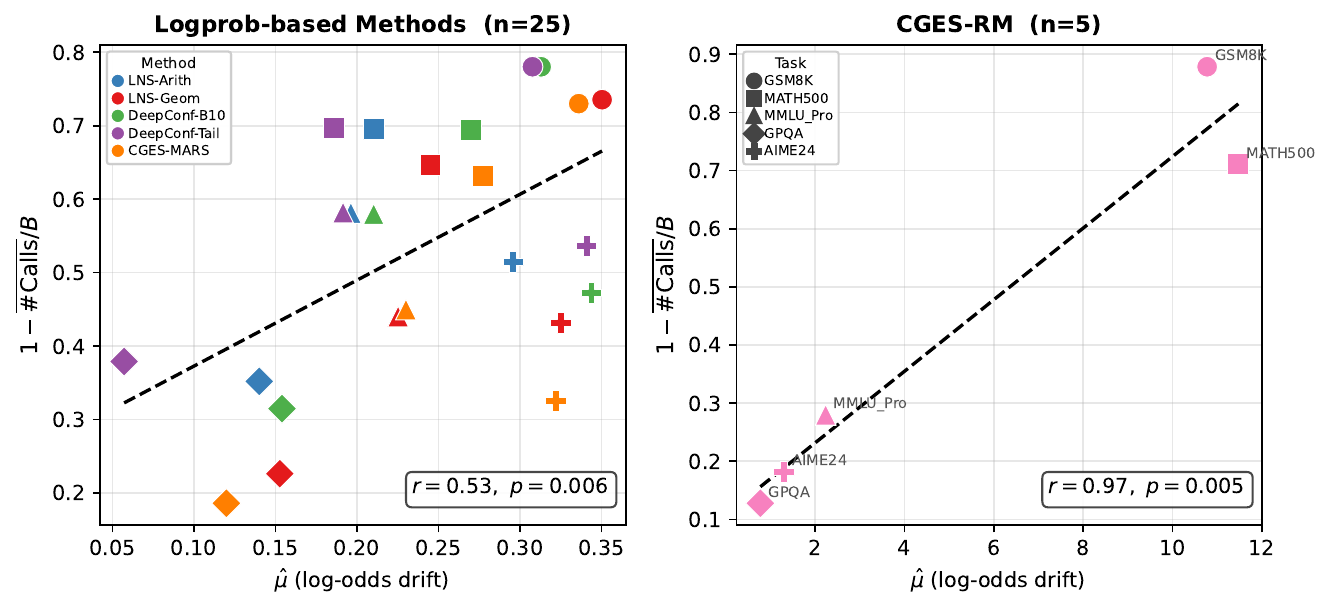}
  \caption{Empirical drift $\hat\mu$ vs.\ CGES efficiency $(1-\overline{\#\text{Calls}}/B)$ at $\gamma{=}0.999$, $B{=}16$. Each point is one (method, task) pair. \textbf{Left}, the five logprob-based methods over 25 pairs ($r{=}0.53$, $p{=}0.006$). \textbf{Right}, CGES-RM over the five tasks ($r{=}0.97$, $p{=}0.005$). Within each family a larger drift yields a larger call reduction, as Theorem~\ref{thm:realistic} predicts. The two panels are kept separate because PRM and token-probability scores live on different scales, so the $x$-axes are not comparable.}
  \label{fig:muhat_efficacy}
\end{figure}

\paragraph{Empirical drift.} The quantity the theorem actually names is the drift. Its empirical estimate is
\[
\begin{aligned}
\hat\mu &\;=\; \mathbb{E}[\ell(C_t)\mid\text{correct}] \;-\; \mathbb{E}[\ell(C_t)\mid\text{incorrect}], \\
\ell(c) &\;=\; \log\tfrac{c(K-1)}{1-c},
\end{aligned}
\]
the sample version of the condition $\mu_k > 0$, where $\ell(c) = \log\frac{c(K-1)}{1-c}$ is the log-odds transform from Algorithm~\ref{alg:cges}. Its sign predicts whether CGES converges and its magnitude predicts how fast. Figure~\ref{fig:muhat_efficacy} plots $\hat\mu$ against CGES efficiency. Within each confidence family the two are strongly correlated, with $r{=}0.97$ ($p{=}0.005$) for CGES-RM across the five tasks and $r{=}0.53$ ($p{=}0.006$) for the logprob-based methods across 25 (method, task) pairs. Pooling all 30 pairs gives a weaker correlation ($r{=}0.30$) because PRM rewards and token-probability scores occupy different ranges. PRM scores spread across $[0,1]$ while token probabilities cluster near $0.9$, and the log-odds transform $\ell$ stretches the former far more than the latter. This is visible in the two panels of Figure~\ref{fig:muhat_efficacy}, where the CGES-RM drift axis runs to about 12 while the logprob axis stays below 0.4. A standardized drift $\hat\mu/\sigma_{\ell(C)}$ partly removes this scale gap, and a fully scale-free cross-family predictor is left to future work. In practice $\hat\mu$ is best read as a task-level predictor. It answers whether CGES will help on a given dataset, which a practitioner can check on a small labelled slice before deploying.

\begin{table}[t]
\centering
\scriptsize
\setlength{\tabcolsep}{3pt}
\caption{Confidence signal quality per method and task. ECE (lower is better calibrated). AUC-ROC (higher discriminates correct from incorrect responses better). Efficiency $=1-\overline{\#\text{Calls}}/B$ at $B{=}16$, $\gamma{=}0.999$. GPQA shows the gap. CGES-RM has low ECE there, comparable to MATH500, yet near-random AUC, which explains its poor efficiency.}
\label{tab:calib_auc}
\resizebox{\linewidth}{!}{%
\begin{tabular}{l|rrr|rrr|rrr|rrr|rrr}
\toprule
Method & \multicolumn{3}{c|}{AIME24} & \multicolumn{3}{c|}{GPQA} & \multicolumn{3}{c|}{MATH500} & \multicolumn{3}{c|}{GSM8K} & \multicolumn{3}{c}{MMLU-Pro} \\
\cmidrule(lr){2-4} \cmidrule(lr){5-7} \cmidrule(lr){8-10} \cmidrule(lr){11-13} \cmidrule(lr){14-16}
& ECE & AUC & Eff. & ECE & AUC & Eff. & ECE & AUC & Eff. & ECE & AUC & Eff. & ECE & AUC & Eff. \\
\midrule
CGES-LNS (arith)    & 0.268 & 0.752 & 0.514 & 0.392 & 0.648 & 0.352 & 0.199 & 0.641 & 0.695 & 0.029 & 0.725 & 0.780 & 0.292 & 0.593 & 0.581 \\
CGES-LNS (geom)     & 0.169 & 0.766 & 0.431 & 0.267 & 0.653 & 0.226 & 0.159 & 0.647 & 0.647 & 0.030 & 0.719 & 0.735 & 0.188 & 0.597 & 0.441 \\
CGES-DeepConf (B10) & 0.219 & 0.824 & 0.472 & 0.360 & 0.677 & 0.315 & 0.199 & 0.673 & 0.694 & 0.029 & 0.725 & 0.780 & 0.291 & 0.599 & 0.580 \\
CGES-DeepConf (tail)& 0.284 & 0.760 & 0.537 & 0.423 & 0.548 & 0.379 & 0.199 & 0.641 & 0.697 & 0.029 & 0.725 & 0.780 & 0.292 & 0.593 & 0.582 \\
CGES-MARS           & 0.117 & 0.826 & 0.326 & 0.205 & 0.650 & 0.186 & 0.139 & 0.673 & 0.632 & 0.035 & 0.720 & 0.730 & 0.188 & 0.617 & 0.450 \\
\midrule
CGES-RM             & 0.216 & 0.682 & 0.181 & 0.050 & 0.685 & 0.128 & 0.050 & 0.954 & 0.711 & 0.021 & 0.940 & 0.879 & 0.141 & 0.758 & 0.280 \\
\bottomrule
\end{tabular}%
}
\end{table}

\section{Robustness to Confidence Noise}
\label{appex:noise}

\begin{figure*}[t]
  \centering
  \includegraphics[width=\linewidth]{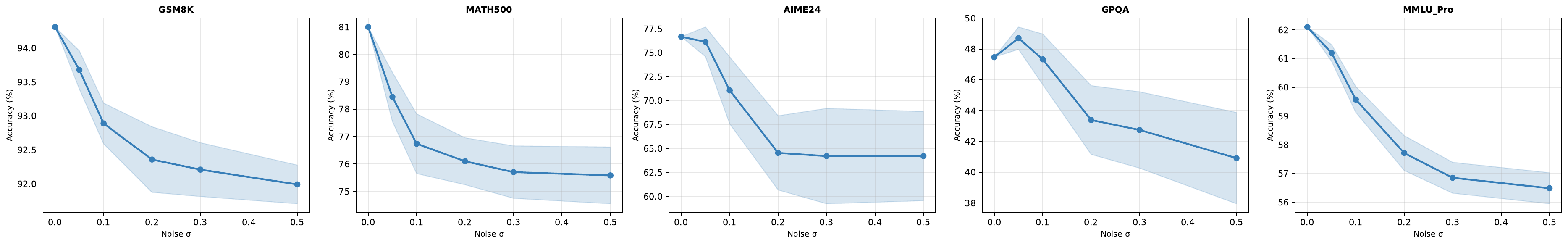}
  \caption{Accuracy of CGES under additive Gaussian noise injected into the confidence scores ($\sigma$ on the $x$-axis). Bands show $\pm 1$ standard deviation over 10 seeds at $B{=}16$. Small $\sigma$ leaves accuracy nearly unchanged. Larger $\sigma$ degrades accuracy gracefully, consistent with the directional-drift consistency result.}
  \label{fig:noise_robustness}
\end{figure*}

To stress-test the directional-drift assumption we add Gaussian noise of standard deviation $\sigma$ to each confidence score (after clipping to $(0,1)$). Figure~\ref{fig:noise_robustness} plots accuracy as a function of $\sigma$ on all five benchmarks. CGES is essentially flat for $\sigma\leq 0.05$ on every task and degrades gracefully for larger $\sigma$. On GSM8K and MATH500 the drop is roughly 2 to 5pp at $\sigma{=}0.5$. On AIME24 the drop is larger because the calibration was already weaker. The shape of these curves matches the theory. Noise shrinks the drift $\mu_k$ but does not flip its sign until $\sigma$ is large enough to dominate the signal.

\section{Selecting the Stopping Threshold $\gamma$ in Practice}
\label{appex:gamma_grid}

A practical question is how to pick $\gamma$ without a labelled validation set. Two options work in our experiments. First, $\gamma$ has a direct calibrated interpretation in the ideal regime (Proposition~\ref{prop:risk_bound}). A user who wants posterior risk at most $1-\gamma$ can pick $\gamma$ directly. Second, in the noisy regime, the empirical drift $\hat\mu$ on a small labelled slice (50 to 100 questions) predicts efficiency within a confidence family (Appendix~\ref{appex:drift}). The user can pick $\gamma$ by sweeping on this slice. In our tables, the efficient $\gamma$ lies between $0.90$ and $0.999$ for every (method, task) pair, and the curves in Figure~\ref{fig:pareto_main} are flat near the top of this range, so the choice is not sharp.

\section{The Use of Large Language Models (LLMs)}
We used large language models (LLMs) solely to aid with polishing the writing and improving clarity of exposition. No part of the research ideation, methodology, analysis, or experimental results was generated by LLMs. The authors take full responsibility for the content of this paper.

\end{document}